\documentclass[journal]{IEEEtran}

\usepackage{ifpdf}
\usepackage{amsmath,amsfonts}
\usepackage{amssymb}  
\usepackage{algorithmic}
\usepackage{algorithm}
\usepackage{array}
\usepackage[caption=false,font=footnotesize,subrefformat=parens,labelformat=parens]{subfig}
\usepackage{textcomp}
\usepackage{stfloats}
\usepackage{url}
\usepackage{verbatim}
\usepackage{graphicx}
\usepackage{cite}
\usepackage{booktabs}
\usepackage[export]{adjustbox}
\usepackage{bm}
\usepackage{nccmath}
\usepackage{mathtools}
\usepackage{hyperref}
\usepackage{tensor}
\usepackage[dvipsnames]{xcolor}
\usepackage{units}
\usepackage{tikz-cd}
\usepackage{bbm}
\usepackage{tikz}
\usepackage{multirow}
\usepackage{accents}

\newcommand{\R}{\mathbb{R}}

\newcommand{\bmat}[1]{\begin{bmatrix} #1 \end{bmatrix}}

\mathchardef\mhyphen="2D

\newcommand{\diag}{\operatorname{diag}}

\newcommand{\CII}{\operatorname{CII}}
\newcommand{\rCII}{\operatorname{rCII}}

\newcommand{\bM}{\mathbf}

\DeclareMathSymbol{\shortminus}{\mathbin}{AMSa}{"39}

\DeclareMathOperator*{\argmax}{argmax}
\DeclareMathOperator*{\argmin}{argmin}

\newcommand{\ubar}[1]{\underaccent{\bar}{#1}}

\begin{document}

\title{Tello Leg: The Study of Design Principles and Metrics for Dynamic Humanoid Robots }
\author{Youngwoo Sim$^1$,~\IEEEmembership{Student,~IEEE,} and Joao Ramos$^1$,~\IEEEmembership{Member,~IEEE,}
\thanks{$^{1}$Authors are with the Department of Mechanical Science and Engineering at the University of Illinois at Urbana-Champaign, Urbana, IL 61801, USA. Corresponding author: {\tt\small sim17@illinois.edu}}
\thanks{{This work is supported by the National Science Foundation via grant CMMI-2043339.}}}


\maketitle

\begin{abstract}
To be useful tools in real scenarios, humanoid robots must realize tasks dynamically. This means that they must be capable of applying substantial forces, rapidly swinging their limbs, and also mitigating impacts that may occur during the motion. Towards creating capable humanoids, this letter presents the leg of the robot TELLO and demonstrates how it embodies two new fundamental design concepts for dynamic legged robots. The limbs follows the principles of: (i) Cooperative Actuation (CA), by combining motors in differential configurations to increase the force capability of the limb. We demonstrate that the CA configuration requires half the motor torque to perform a jump in comparison to conventional serial design configurations. And (ii) proximal actuation, by placing heavy motors near the body to reduce the inertia of the limb. To quantify the effect of motor placement on the robot's dynamics, we introduce a novel metric entitle Centroidal Inertia Isotropy (CII). We show that the design of state-of-the-art dynamic legged robots empirically increase the CII to improve agility and facilitate model-based control. We hope this metric will enable a quantifiable way to design these machines in the future. 

\end{abstract}


\IEEEpeerreviewmaketitle

\section{introduction}

\IEEEPARstart{C}{apable} humanoid robots could be directly inserted in environments designed for humans without major structural modifications. These machines could help firefighters in emergency response; could help nurses transport weakened patients; or help workers move heavy boxes within warehouses. However, to be useful tools capable of realizing physical tasks with human-level performance, humanoid robots must be \textit{dynamic}. Dynamic whole-body movements enable robots to amplify the forces that they can apply to the environment. For instance, the fast movement performed by weightlifters allows them to raise a heavy barbell over their heads, a feat that would be impossible to realize semi-statically. Designing robots that realize tasks dynamically enables minimizing the robot size and mass, making them safer to interact with humans and more cost effective. 

\begin{figure}[!t]
\centering
\adjincludegraphics[trim={0 {0.33\height} {0.47\width} 0},clip,width = 1\linewidth]{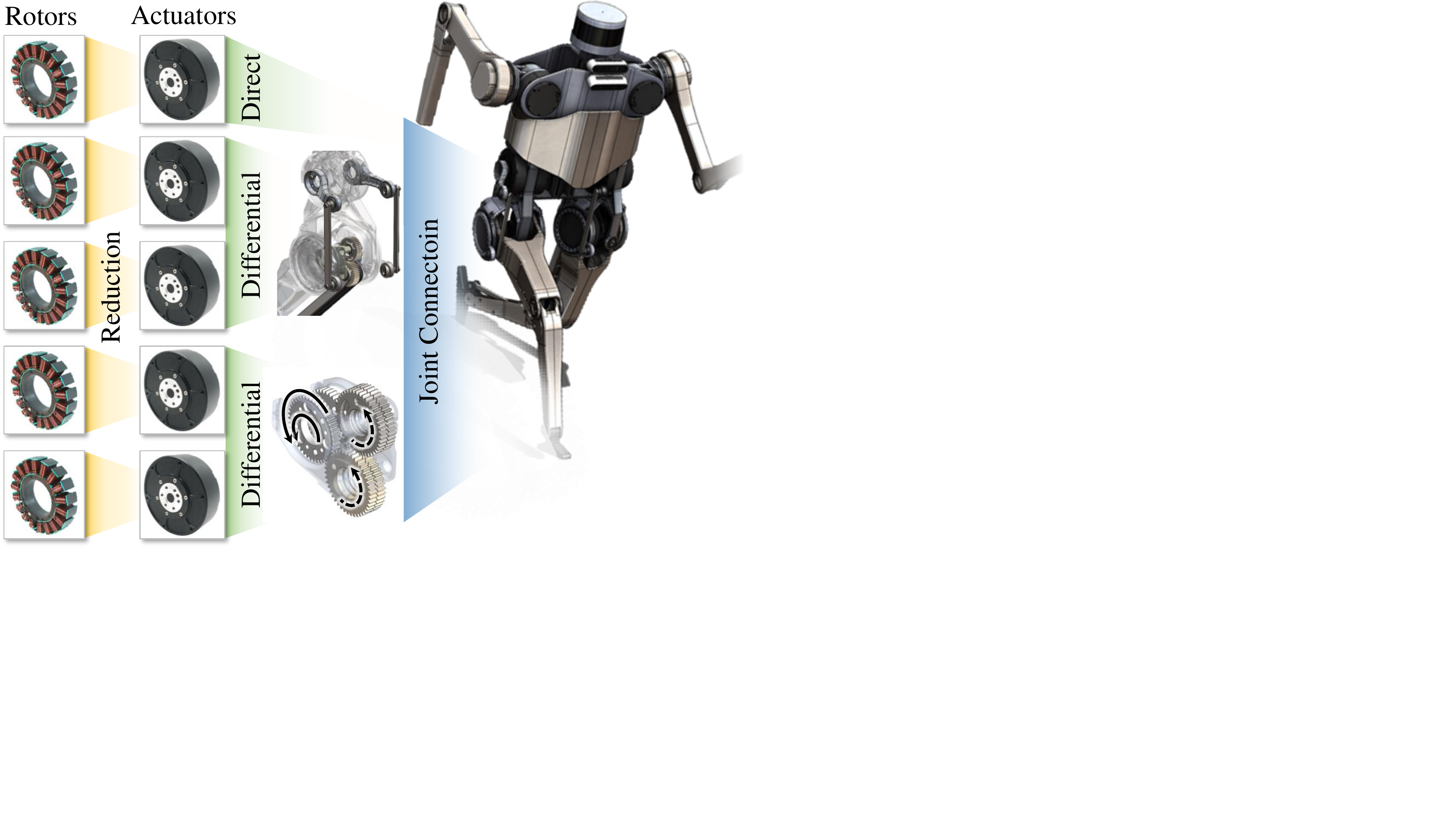}
\caption{Physical layers of actuation system in the leg of the humanoid Tello. Motors are arranged in differential configurations to increase the limb's force capability without substantially increasing reflected inertia.}
\label{FirstFig}
\end{figure}

Realizing dynamic tasks require operation at the boundaries of the robot's performance. But extending these boundaries requires the formalization of \textit{principles} that guide the design of candidate hardware and \textit{metrics} that enable the quantification of their performance. Towards this goal, this work introduces a novel design principle entitled \textit{Cooperative Actuation (CA)} and a novel performance metric we name \textit{Centroidal Inertia Isotropy (CII)} that are tailored to dynamic legged robots. These concepts were employed in the design of the leg of the dynamic humanoid Tello. Here we show experimental results that demonstrate their use.

On one hand, \textit{design principles} are rules that guide the process of conceiving the hardware of the robot. For instance, Seok demonstrated that employing electric actuators with large gap radius and low gearing improve running efficiency of legged robots \cite{seok2012actuator}. Kim \cite{LIMS} built backdrivable cable-driven manipulators that are safe for physical human robot interaction by reducing contact inertia. Sim \cite{sim2021dynamic} and Kikuwe \cite{kikuuwe2022dynamics} showed that employing energy efficient mechanical transmissions reduces the reflected inertia of limbs and improve backdrivability. Similarly, the CA is a design principle for legged robots for decreasing the torque requirement per motor without inflating the reflected inertia of the limb. This is achieved by combining multiple motors to simultaneously drive multiple joints of the robot's limb that are often not active at the same time. This is in contrast with the common approach of increasing the gearing ratio of transmissions in order to increase the output torque of the actuators, which quadratically increases its reflected inertia. The Tello leg employs a pair of actuators to {drive the knee and ankle joints}, and a second pair that together {drive hip abduction and flexion.}

On the other hand, \textit{performance metrics} quantify the robot's ability to meet a task requirement. These metrics also work as a common language to compare the performance of different machines. For instance, 
Wensing \cite{ImpactMitigation} introduced the Impact Mitigation Factor (IMF) that quantifies the energetic losses due to inertial loads when the robot's limb impacts the ground.
Chiacchio \cite{chiacchio2000manipulability} demonstrated that actuators' capability as polytopes, as depicted in Fig. \ref{fig:CaseStudyCA}, represents the exact torque and velocity capability for a given limb configuration. Jeong \cite{jeong2009impact} introduced the dynamic acceleration polytope, which represents maximum achievable acceleration in task space respecting actuator limits. This work introduces the CII as a metric for estimating how closely the robot's full dynamics can be approximated by its centroidal dynamics model \cite{orin2008centroidal}. This reduced model is widely used for planning and control of highly dynamic behaviors for legged robots due to its simplicity for fast computation and competence to capture the core dynamics of the full robot 
\cite{apgar2018fastCassie, cassieHLIP, gibson2021terrainALIP, huboPlusDyn,JaredSRB_MPC, GerardoRPC,multiContactCaron, GarciaCentroidalMPC}. For instance, in \cite{kim2019highly} the Mini Cheetah performs agile running using a combination of a model predictive controller based on the centroidal model with a cascaded whole-body controller. Because the robot's limbs are extremely lightweight, the centroidal model can accurately represent the full robot dynamics, allowing WBC to easily map the MPC trajectory to the full robot at high control rates around 1kHz. Although the centroidal model has also been extensively used for planning and control of  dynamic humanoid robots \cite{multiContactCaron, GarciaCentroidalMPC}, 
it is not as accurate to represent the behavior of these robots due to substantial mass of their limbs. But how lightweight should the limbs of humanoid robots be so the centroidal model can accurately describe the robot's full dynamics? To answer this question, we introduce the CII as a condensed metric to quantify the error between the centroidal model and the full dynamics of the robot. To maximize the CII of the humanoid Tello, all actuators are placed near the body to reduce the moving inertia of the limbs.

The fundamental contributions of this letter are the design principle of CA and the performance metric CII tailored to dynamic humanoid robots. We demonstrate these concepts with the design of the leg of the humanoid Tello. In section \ref{sec:CA}, we discuss a design principle for improving force density and reducing reflected inertia of legged robots by analyzing different configuration of actuators to joint connection. In section \ref{sec:CII}, a performance metric, CII, that quantifies the proximodistal distribution of mass in legged robot is introduced. We demonstrate that the CII can be evaluated for different types of legged robots and eventually classify the robots. In section \ref{sec:TelloImplementation}, a 5 degree-of-freedom (DoF) humanoid leg, Tello, is presented as an embodiment of CA to improve joint torque. The 5 actuators of this leg are placed adjacent to the torso such that distal mass is dramatically reduced. In section \ref{sec:experiments}, an experiment of Tello jumping is analyzed to demonstrate the high joint torque enabled by CA.

\section{Cooperative Actuation} \label{sec:CA}
To realize dynamic tasks, the robot's actuation system needs to deliver large forces when compared to the machine's total mass. In other words, the robot's limbs need to have \textit{high force density}. A straightforward approach to increase the torque limits $\tau_\psi$ of actuators is to increase the gearing ratio $N$ of the transmission ($\tau_\psi\propto N$). However, increasing the ratio $N$ increases friction in the joints and quadratically increases the reflected inertia of the rotors ($I_\textrm{ref}  \propto  N^2$), which degrades the limb's backdrivability and its ability to regulate contact and mitigate impacts \cite{sim2021dynamic,ImpactMitigation}. Hence, gearing ratio cannot be increased indefinitely because backdrivability degrades faster than the increase in torque (force) density. In contrast, \textit{an alternative solution to increase the force density of the limb is to combine multiple motors in parallel to simultaneously drive multiple joints}. This increases the joint torque proportionally to the number of motors while just linearly increasing the reflected inertia of the joint. This approach is especially beneficial when these joints are not often recruited simultaneously. Here we study the design trade-offs of combining multiple actuators in differential configurations to drive multiple joints.


To illustrate the benefit of combining multiple actuators, we study the effect of the \textit{topology Jacobian} which represents the connectivity between actuators and joints. For serial manipulators, because each joint is driven by a single actuator, this Jacobian $\tensor*[^q]{\bM J}{_{\psi,d}}$ is simply given by the diagonal matrix whose entries are reciprocals of gear ratio.
However, when combining two actuators in a \textit{differential} configuration to drive a pair of joint, each joint is always driven by both actuators:
\begin{IEEEeqnarray}{rCl}
    q_1 &=& \dfrac{1}{2N_d} (\psi_1 +  \psi_2 ), \\
    q_2 &=& \dfrac{1}{2N_d} (\psi_1 -  \psi_2 ),
\end{IEEEeqnarray}
where $N_d$ is the gear ratio for differential configuration, $q_1$, $q_2$ are joint displacements and $\psi_1$, $\psi_2$ are actuator displacements. Hence, the mapping from actuator velocity $\dot{\bm \psi}$ to joint velocity  $\dot{\bm q}$ called topology Jacobian $\tensor*[^q]{\bM J}{_\phi}: \dot{\bm \psi} \mapsto \dot{\bm q}$ of differential and serial configurations are, 
\begin{equation}
    \tensor*[^q]{\bM J}{_{\psi,d}} 
    =
    \bmat{\dfrac{1}{2N_d} & \phantom{-}\dfrac{1}{2N_d} \\ \dfrac{1}{2N_d} & -\dfrac{1}{2N_d}}, 
    \quad 
    \tensor*[^q]{\bM J}{_{\psi,s}} 
    =
    \bmat{\dfrac{1}{N_s} & 0 \\ 0 & \dfrac{1}{N_s}},\label{differentialActuationTopology}
\end{equation}
where $N_s$ is the gear ratio of actuators in serial configuration. Unlike velocities' transformation, torques of actuators ${\bm \tau}_\psi$ are converted to joint torques ${\bm \tau}_q$ by inverse of dual maps $\tensor*[^q]{\bM J}{_\psi^{\shortminus\top}}: {\bm \tau}_\psi \mapsto {\bm \tau}_q$, which gives torque conversion Jacobians for differential and serial configurations as follows, 
\begin{equation}
    \tensor*[^q]{\bM J}{_{\psi,d}^{\shortminus\top}}
    =
    \bmat{N_d & \phantom{-}N_d \\ N_d & - N_d}, 
    \quad 
    \tensor*[^q]{\bM J}{_{\psi,s}^{\shortminus\top}} 
    =
    \bmat{N_s & 0 \\ 0 & {N_s}}.\label{differentialActuationTopologyDual}
\end{equation}

\begin{figure}[!t]
  \centering
  \adjincludegraphics[trim={{0.0\width} {.47\height} {0.645\width} {0.0\height}},clip,width =1.0\linewidth]{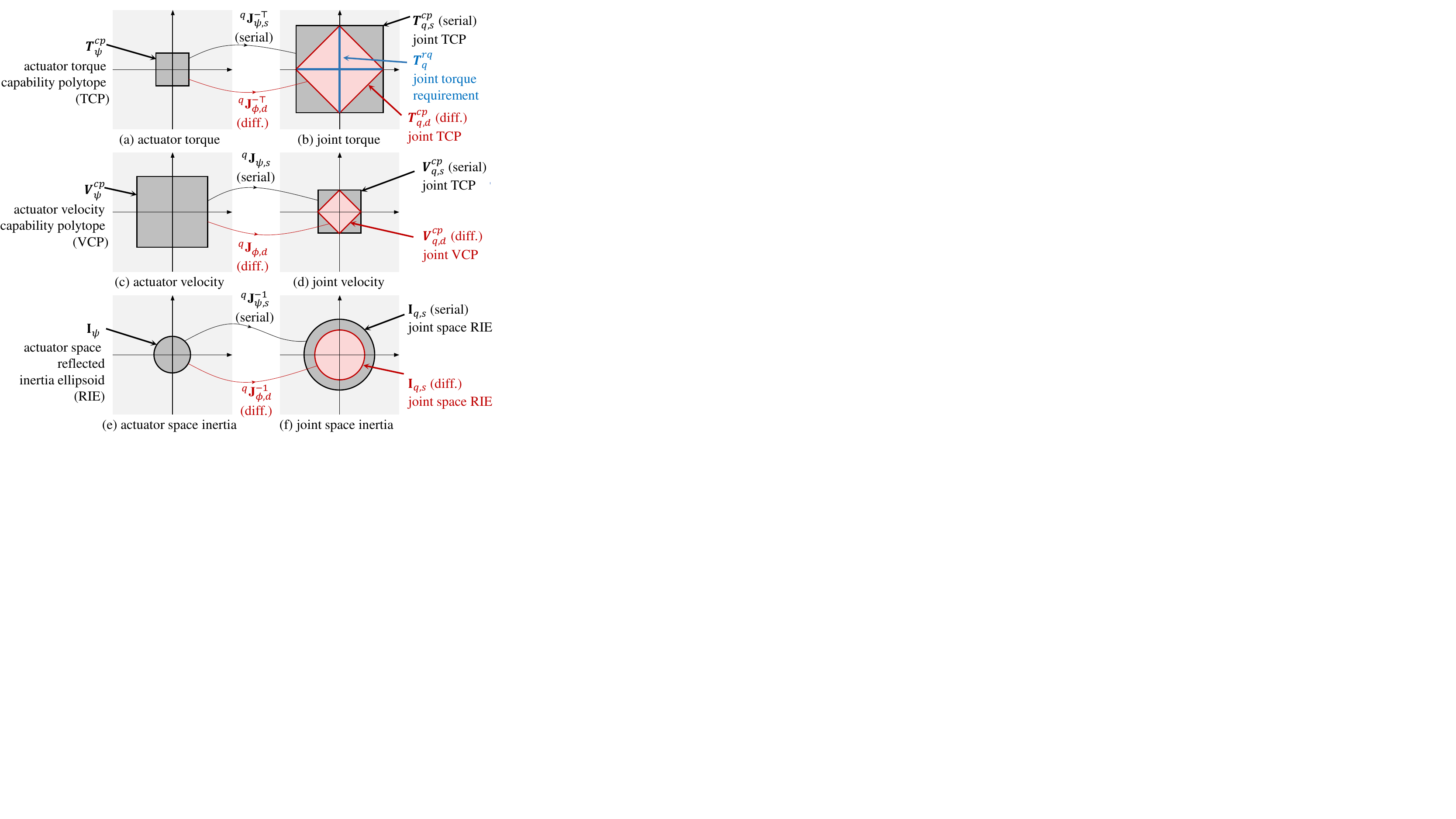}
  \caption{  Torque capability polytopes (top row) and velocity capability polytopes (center row) of actuators are projected onto joint space. The reflected inertia of the rotors of actuators are depicted as ellipse in actuators space and joint space (bottom row). }
  \label{fig:CaseStudyCA}
\end{figure}

The effect of Jacobians can be visually understood by introducing capability polytopes.
The torque and velocity limits of individual actuators are collected as vectors of torque and velocity to form a torque or a velocity capability polytope (TCP, VCP). In serial configuration, the actuator TCP ${\bm T}_\psi^{cp}$ is scaled up to joint TCP ${\bm T}_{q,s}^{cp}$ by the factor of gear ratio $N_s$ which is described as grey boxes in Fig. \ref{fig:CaseStudyCA}(a)(b). In contrast, with differential configuration, the actuator TCP is not only scaled by $N_d$, but also \textit{rotated} 45 degrees to yield joint TCP ${\bm T}_{q,d}^{cp}$ which is described as a red box in Fig. \ref{fig:CaseStudyCA}(b).


The differential configuration can reduce reflected inertia if torques are expected to be applied on a single joint at a time. For example, the knee and ankle torque in bipedal gait do not peak simultaneously, but the concentration of joint torque slide from knee to ankle \cite{PATRIARCO1981513, VAUGHAN1996423, act3010001, huang2001planning, felis2015gait}. Such joint torque usage is roughly modeled as ${\bm T}_q^{rq}$ as blue lines in Fig. \ref{fig:CaseStudyCA}(b). For differential configuration to cover the ${\bm T}_q^{rq}$, the polytope is required to inflate \textit{less} than serial configuration because it rotates to the ${\bm T}_q^{rq}$ more efficienctly. Hence, the gear ratio required differential configuration is the half of gear ratio for serial configuration {to achieve the same required joint torque},  
\begin{IEEEeqnarray}{rCl}
 N_d = \tfrac12 N_s. 
 \label{eqn:gearRatioCA}
\end{IEEEeqnarray}
Consequently, the joint space reflected inertia is smaller in case of differential configuration as in Fig. \ref{fig:CaseStudyCA}(f). The reflected inertia on the join space $\tensor*[^q]{\bM I}{_\psi}$ is given as,
\begin{IEEEeqnarray}{rCl}
    \tensor*[^q]{\bM I}{_\psi}
    &=& 
    \tensor*[^q]{\bM{J}}{_\psi^{\shortminus \top}} 
    \tensor*[^\psi]{\bM I}{_\psi}   
    \tensor*[^q]{\bM{J}}{_\psi^{\shortminus 1}},
    \label{eqn:ReflectedInertia}
\end{IEEEeqnarray}
where $\tensor*[^\psi]{\bM I}{_\psi} 
$ is the rotor inertia matrix whose entries are individual inertia of rotors. 
The joint space reflected inertia in differential $\tensor*[^q]{\bM I}{_{\phi,d}}$ and serial configuration  $\tensor*[^q]{\bM I}{_{\phi,s}}$ are, 
\begin{equation}
    \tensor*[^q]{\bM I}{_{\phi,d}} = \bmat{2N_0^2 & 0 \\ 0 & 2N_0^2},
    \quad 
    \tensor*[^q]{\bM I}{_{\phi,s}} = \bmat{4N_0^2 & 0 \\ 0 & 4N_0^2},
\end{equation}
where $N_0 \coloneqq N_c^* = \tfrac{1}{2}N_s^*$ is used to conveniently compare both inertia tensors. This result indicates that the principal components of reflected inertia of serial configuration are twice larger than those in differential configuration.

The differential configuration also affects velocity conversion between joints and actuators such that there are cases where actuators need to run faster than the joints. In other words, when velocity capability of actuators are given identical to both configurations, differential configuration renders smaller velocity capability on the joint space as described in the conversion of VCP, ${\bm V}_\psi^{cp}$ to ${\bm V}_{q,d}^{cp}$ as in Fig. \ref{fig:CaseStudyCA}(c)(d). 

\subsection{Design Principle: Cooperative Actuation}
To develop an actuation system such that generates sufficiently large force and minimizes reflected inertia, two design considerations are joined together as a design principle called \textit{Cooperative Actuation}. First, it is assumed that joint torques are maximally recruited only on a single joint at a time. In other words, the torques of multiple joints should not peak simultaneously. 
Second, a combination of actuators can work in parallel to drive multiple joints. In this way, load on a joint can be shared by multiple actuators. For example, differential configuration combines actuators' torques so that twice of maximum actuator torque can be applied on a joint. The differential actuation is advantageous under previous assumption because it yields smaller reflected inertia. For serial configuration to yield the maximum joint torque attained by differential configuration, the gear ratio for serial configuration has to be the double of gear ratio for differential configuration as in equation \eqref{eqn:gearRatioCA}. 
Such gear ratios yield joint space reflected inertia of serial configuration twice larger than the joint space reflected inertia of differential configuration. 

\section{Centroidal Inertia Isotropy} \label{sec:CII}

\begin{figure*}[!t]
    \centering
    \subfloat[]{
        \adjincludegraphics[trim={{0.0\width} 0 {0.58\width} 0},clip,height = .19\linewidth]{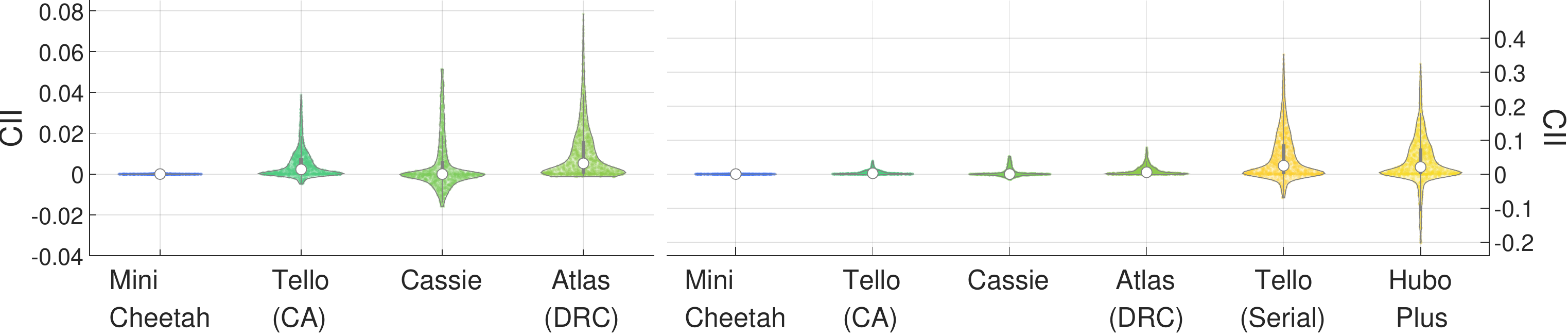}
        \label{fig:CiiDistProximal}
    }
    \quad
    \subfloat[]{
        \adjincludegraphics[trim={{0.42\width} 0 {0.0\width} 0},clip,height = .19\linewidth]{Fig/RAL2022_Fig_CII_Distribution.pdf}
        \label{fig:CiiDistAll}
    }
    \caption{The CII of various robots are evaluated for $30\times 30$ configurations in equi-distance grid and their distribution is demonstrated. (a) Within the robots employing proximal actuation, the order of the range of CII differs between a quadruped and humanoids. (b) A class of robots employing proximal actuation and serial robots whose actuators are placed adjacent to their joint are compared. }
    \label{fig:CiiDistribution}
\end{figure*}

The controllers of dynamic legged robots such as Cassie, MIT Mini Cheetah, or Atlas employ a planners based on reduced models such as variations of the linear inverted pendulum (LIP) models \cite{apgar2018fastCassie, cassieHLIP, gibson2021terrainALIP} spring loaded inverted pendulum (SLIP) model \cite{huboPlusDyn}, single rigid body model (SRBM) \cite{JaredSRB_MPC, GerardoRPC} or centroidal dynamics \cite{multiContactCaron, GarciaCentroidalMPC}. However, the actual dynamics of legged robots differs from these simple models, because the mass of the limbs contributes to inertia of the whole system. To reduce this difference, aforementioned robots are built with lightweight limbs to resemble simpler models. For example, actuators, which contributes to large part of the total moving mass, are placed near the center of mass or torso. {\textit{This design principle has been empirically used in the design of dynamic legged robots, but has never been quantified.}} The premise of this approach, called proximal actuation, is to minimize swing mass and inertia of the limb to facilitate control and improve agility. 

Although the proximodistal distribution of robot's mass ultimately limits the frequency at the planning runs and require controllers based on more complex models \cite{posaTO, mordatch2015ensemble, daiCentroidalWBC, herzogKinoDyn, kim2019highly, wang2021unified}, the relation between mass distribution and performance of controllers is still poorly studied. Hence, it is difficult to draw a line as to how much proximal actuation is needed to attain control objective. In case of humanoid robots, it is more challenging to achieve proximal actuation. First, placing more than 10 electrical motors around hip requires extensive space optimization and more stages of transmission that delivers actuator outputs to distal joints. Moreover, the inertia of the structural bodies are larger than quadrupeds due to longer limb lengths. 

This section introduces a design metric, called \textit{Centroidal Inertia Isotropy}, to quantify the state of proximodistal mass distribution as a caliber of proximal actuation. The foundational logic behind this metric is that the mass distribution is not directly measured, but the change in inertia of the whole system due to pose (joint configuration) change is quantified, which gives the name inertia isotropy.   
For example, if a robot has massless limbs, the inertia of the whole system will remain the same regardless of joint configuration. In other words, the inertia is perfectly isotropic on the joint space. In contrast, the more the mass is distributed toward the distal ends of limbs, the more pronounced is the difference in the inertia of the whole system at different body configurations.
The concept of distance between two different inertia matrices is formulated with centroidal dynamics.  
Consider a robot at nominal configuration ${\bm q}_0$ with centroidal angular momentum ${\bm h}_G \in \mathbb{R}^3$ in task space, 
\begin{equation}
    {\bm h}_{G} = {\bM I}_{G} ({\bm q}_0) {\mathbf{v}}_0,
\end{equation}
where ${\bM I}_G\in \mathbb{R}^{3\times3}$ is a submatrix of the original $6\times6$ CCRBI matrix in \cite{orin2008centroidal} which accounts for the (rotational) inertia, and ${\mathbf{v}_0} \in \mathbb{R}^3$ is a generalized angular velocity vector in configuration ${\bm q}_0$. 
Note that subscript $G$ associates its parent vector to a coordinate frame at the point of center of mass in task-space.

Next, consider the robot changing its joint configuration to $\bm q$ without external wrench. The CCRBI in configuration $\bm q$ is derived using the principle of momentum conservation,
\begin{equation}
     {\bm h}_{G} 
     =
     {\bM I}_{G} ({\bm q}_0) {\mathbf{v}}_0 
     =
     {\bM I}_{G} ({\bm q}) {\mathbf{v}}. 
\end{equation}
If limbs were massless and it is only the torso contributing to CCRBI, both CCRBI and velocities would remain unchanged after any change in configuration. However, if limbs are \textit{not} massless, the change from ${\bm q}_0$ to $\bm q$ leads to difference in generalized velocities, ${\mathbf{v}}$ and ${\mathbf{v}}_0$, 

\begin{equation}
    \mathbf{v} - {\mathbf{v}}_0 
    = ( 
    {\bM I}_G (\bm q)  
    {\bM I}^{\shortminus1}_G ({\bm q}_0) 
    - 
    \mathbf{1}_3
    ) {\mathbf{v}}_0,
\end{equation}
where $\mathbf{1}_3 \in \R^{3\times 3}$ is the identity matrix.
After normalizing the difference in the velocities by the generalized velocity at nominal configuration ${\mathbf{v}}_0$ and taking the determinant of the coefficient matrix, the CII is obtained,
\begin{equation}
    \CII \big(\bm q, {\bm q}_0 \big) 
    :=
    \: \det
    \big(  
        \tensor*[]{\bM I}{_G} (\bm q)
        {\bM I}^{\shortminus1}_G ({\bm q}_0)
        - 
        \mathbf{1}_3
    \big) .
\end{equation}

\begin{figure}[!t]
    \begin{center}
    \small
    \label{tbl1}
    \centering
    \begin{tabular}{@{}cc@{}@{}c@{}@{}c@{}@{}c@{}@{}c@{}@{}c@{}}
      \toprule
        & Mini & \textbf{Tello} & \multirow{2}{*}{Cassie} & Atlas & Tello  & Hubo\\
        & Cheetah & \textbf{(CA)} &  & (DRC) & (Serial) & Plus\\
        \specialrule{.4pt}{2pt}{3pt}
        \rotatebox{90}{Max. CII}
        &
        \adjincludegraphics[trim={{0.2\width} {0.25\width} {0.15\width} {0.1\width}},clip,width = .15\linewidth,valign=b]{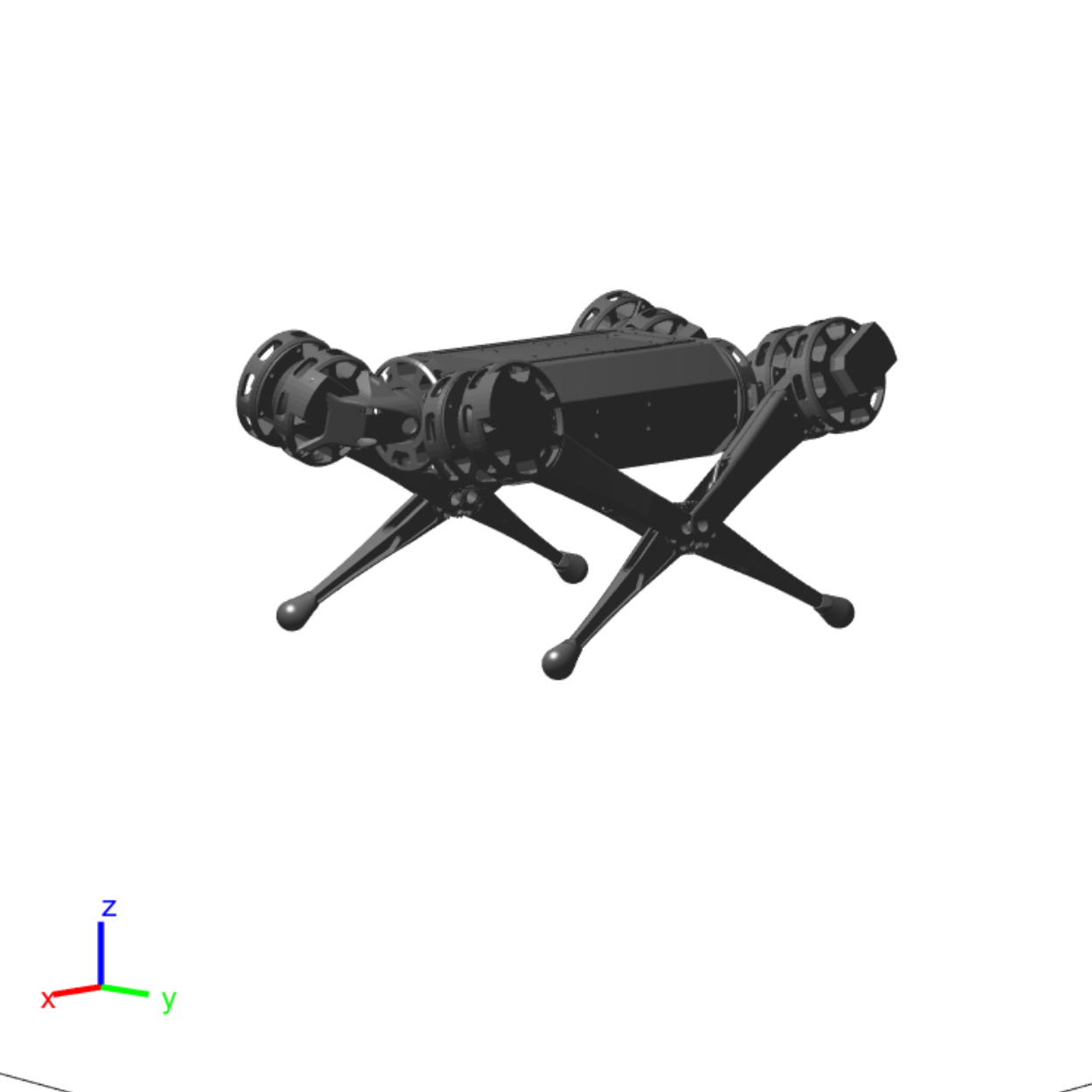}  
        &
        \adjincludegraphics[trim={{0.22\width} {0.20\width} {0.22\width} {0.24\width}},clip,width = .15\linewidth,valign=b]{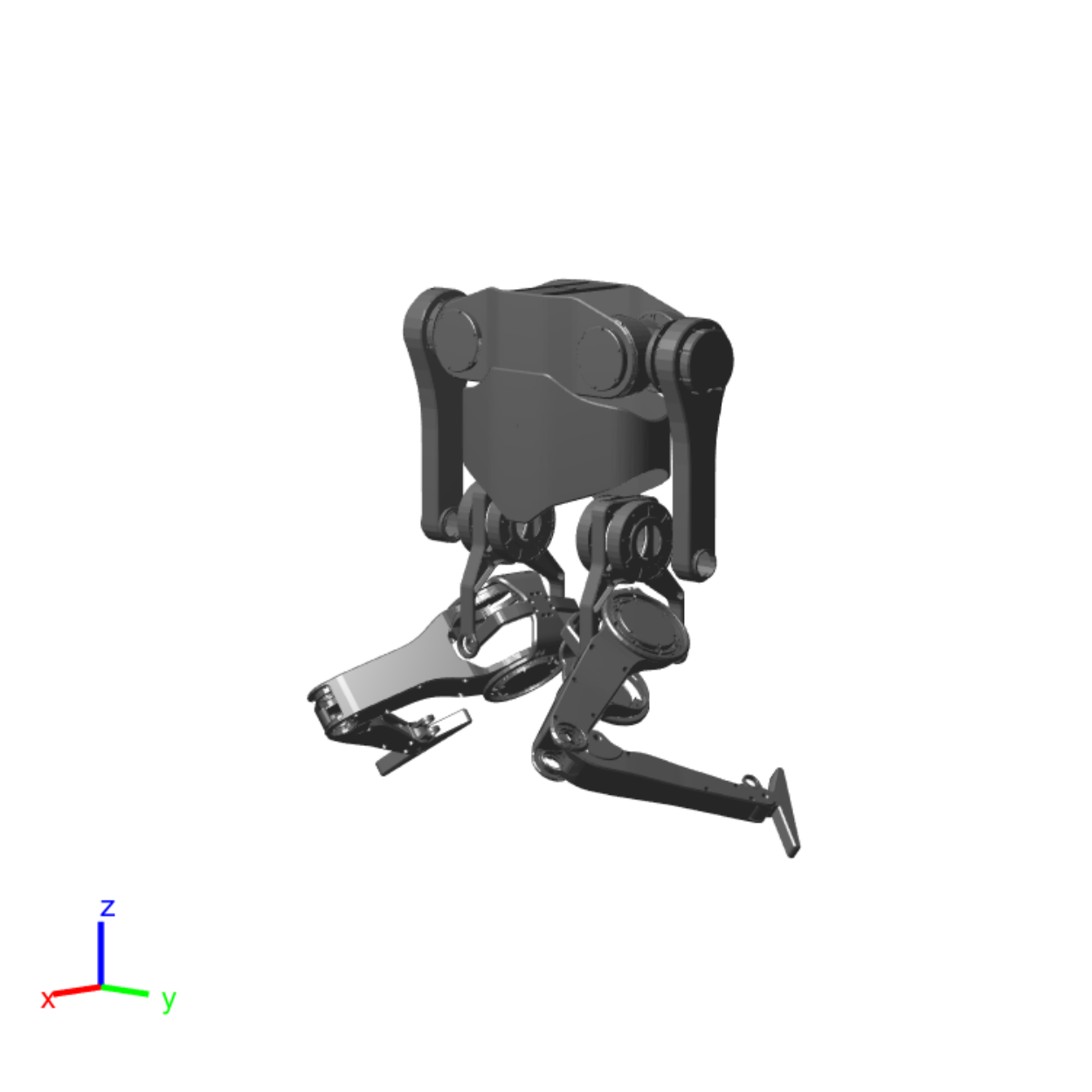} 
        &
        \adjincludegraphics[trim={{0.14\width} {0.16\width} {0.14\width} {0.12\width}},clip,width = .15\linewidth,valign=b]{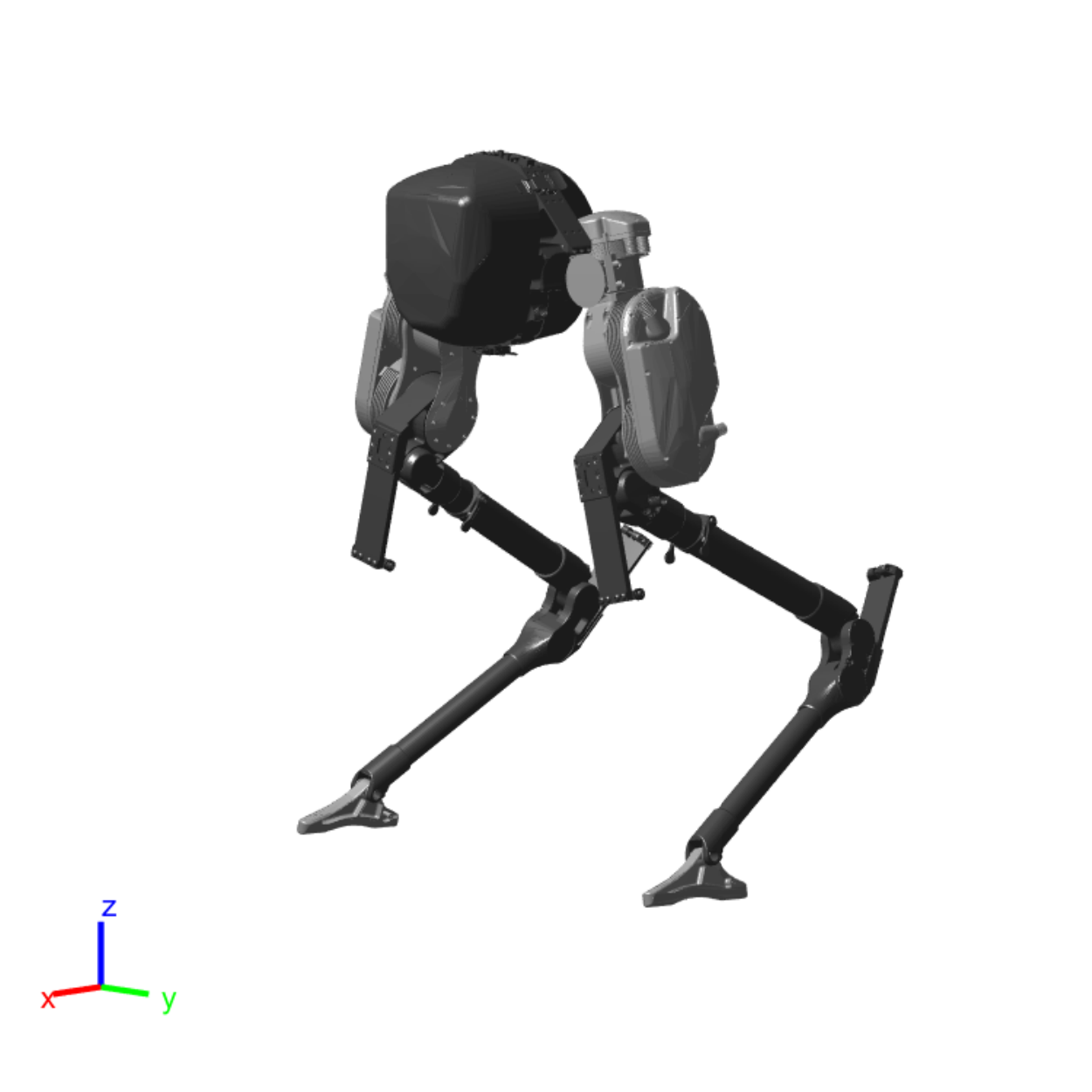} 
        &
        \adjincludegraphics[trim={{0.15\width} {0.17\width} {0.15\width} {0.13\width}},clip,width = .15\linewidth,valign=b]{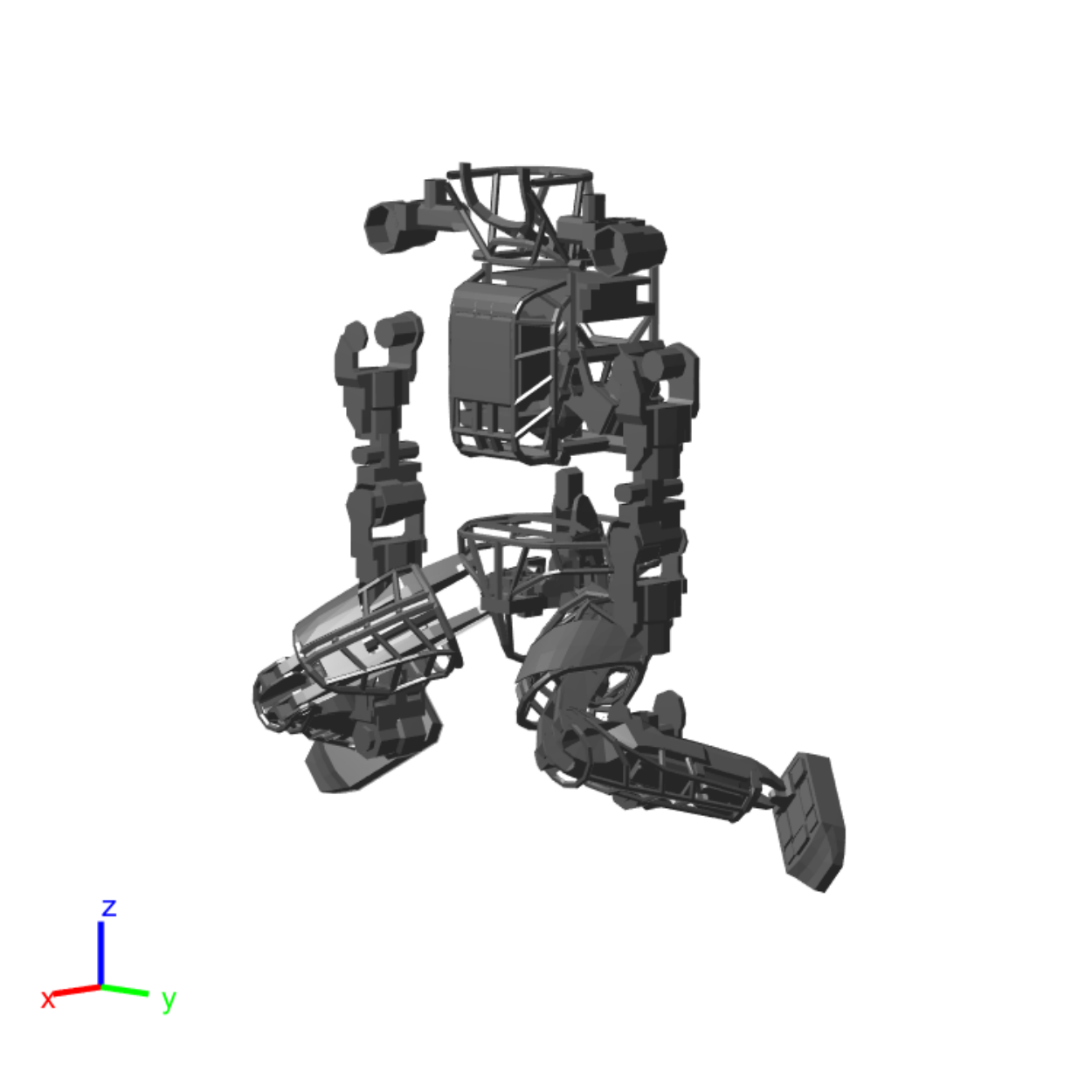} 
        &
        \adjincludegraphics[trim={{0.22\width} {0.20\width} {0.22\width} {0.24\width}},clip,width = .15\linewidth,valign=b]{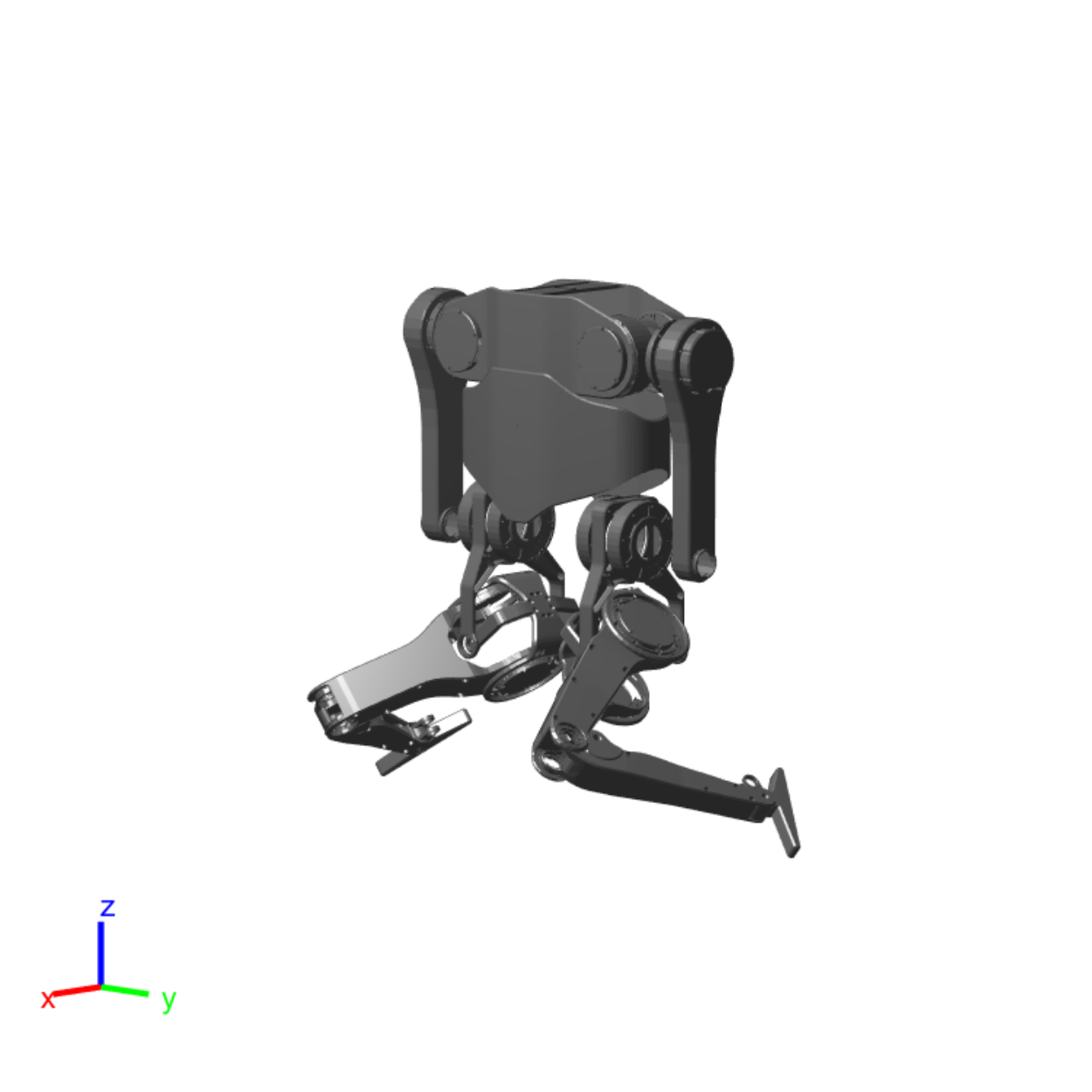} 
        &
        \adjincludegraphics[trim={{0.17\width} {0.17\width} {0.17\width} {0.17\width}},clip,width = .15\linewidth,valign=b]{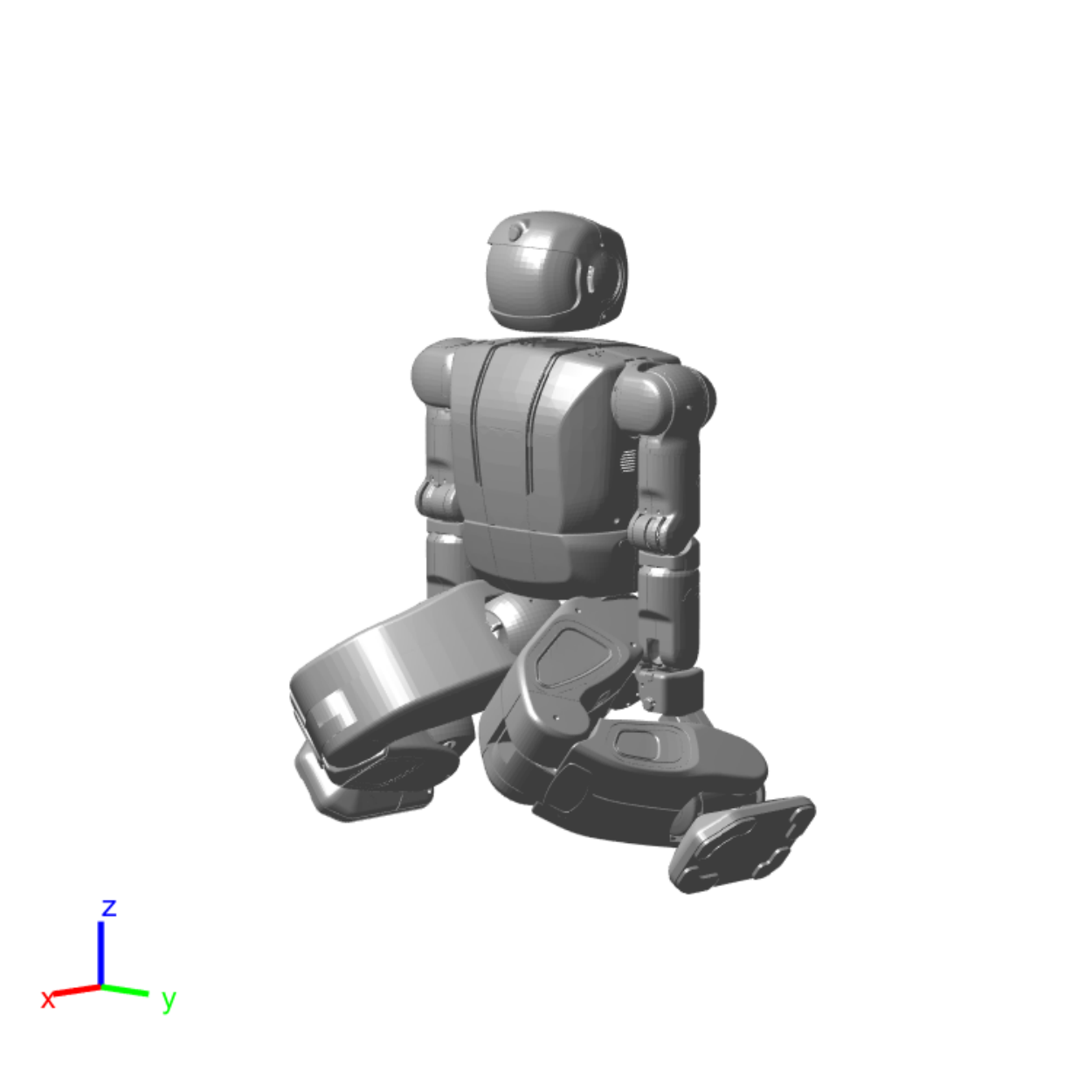} 
        \\
        \specialrule{.4pt}{3pt}{3pt}
        \rotatebox{90}{Min. CII}
        &
        \adjincludegraphics[trim={{0.18\width} {0.2\width} {0.12\width} {0.1\width}},clip,width = .15\linewidth,valign=b]{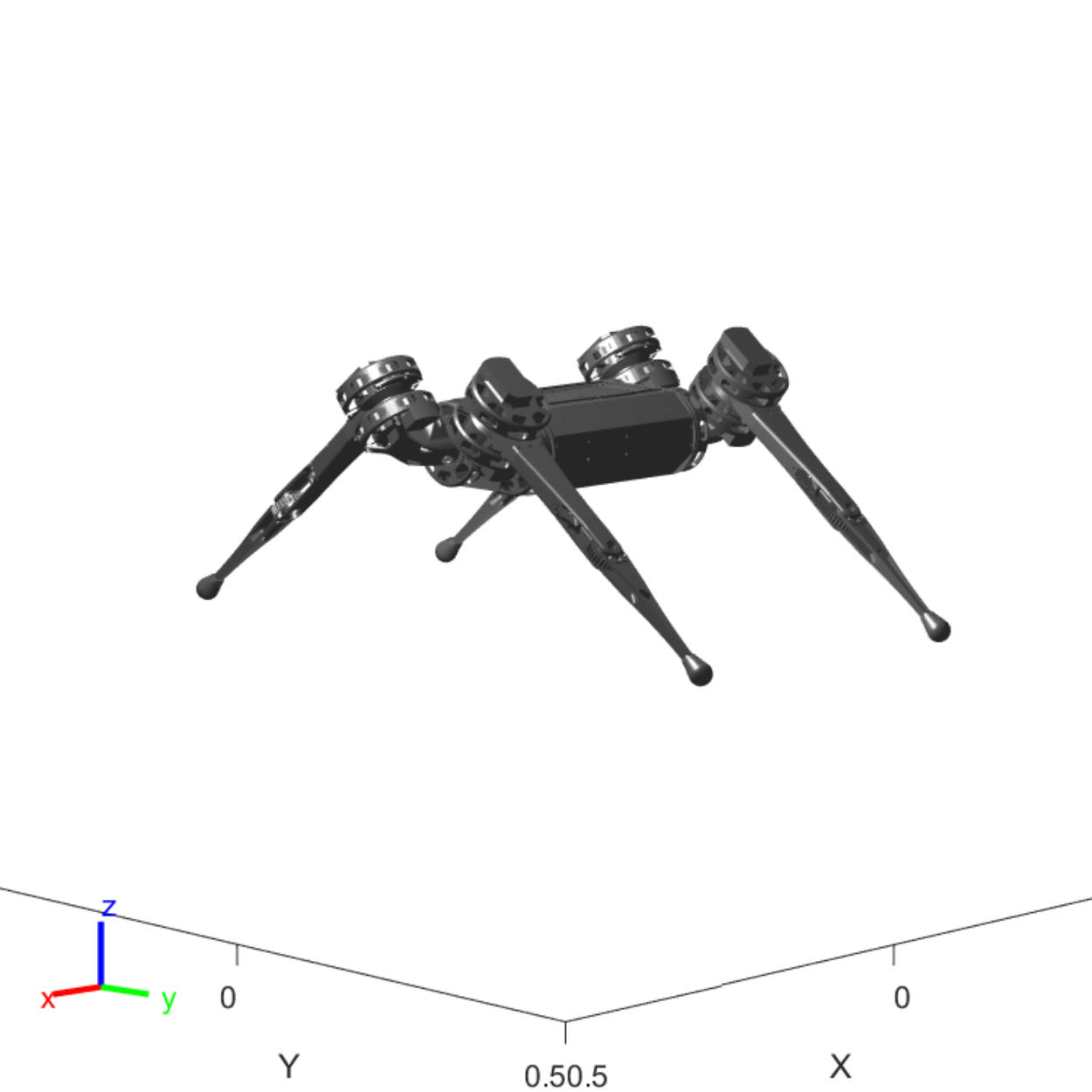}  
        &
        \adjincludegraphics[trim={{0.23\width} {0.15\width} {0.17\width} {0.25\width}},clip,width = .15\linewidth,valign=b]{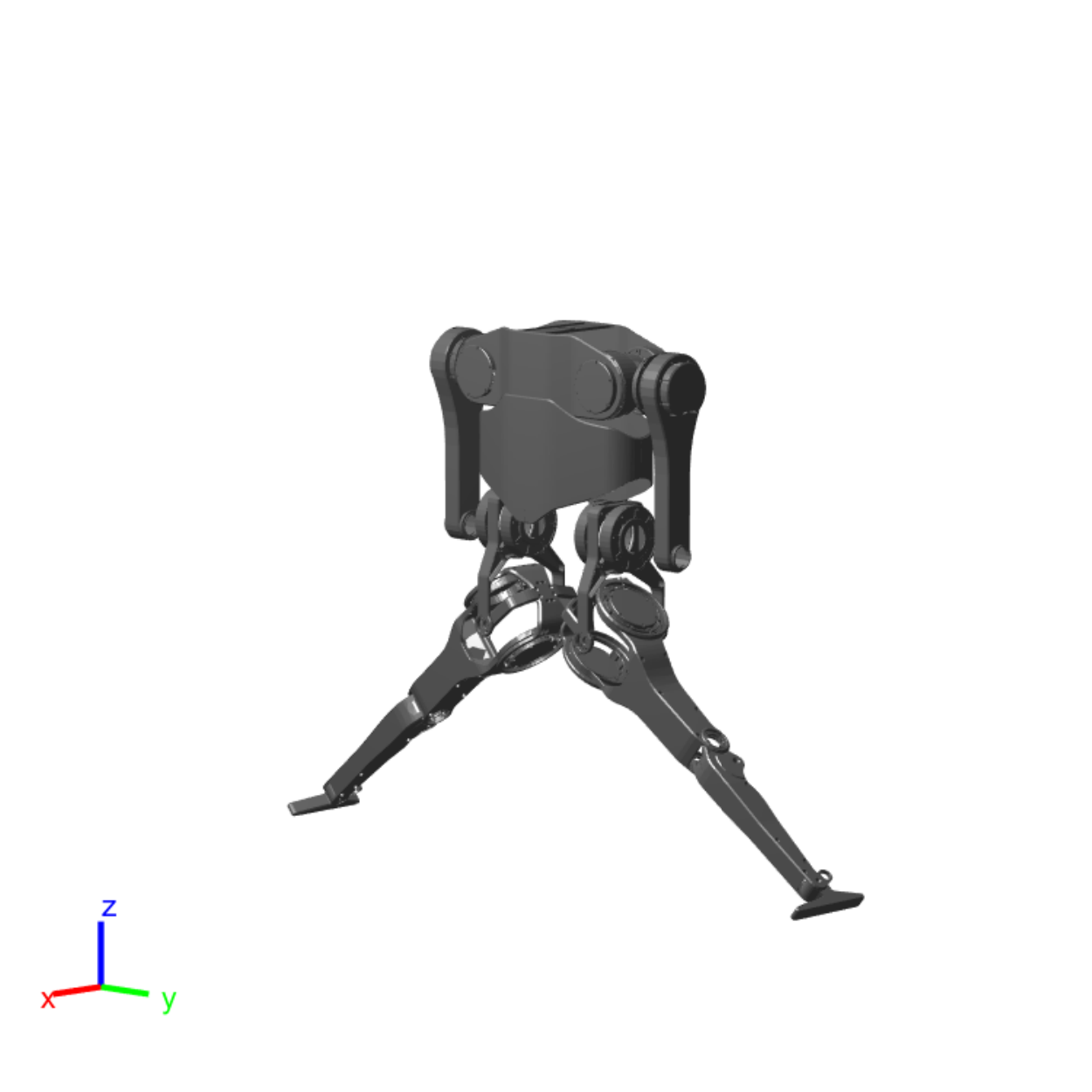} 
        &
        \adjincludegraphics[trim={{0.12\width} {0.12\width} {0.12\width} {0.12\width}},clip,width = .15\linewidth,valign=b]{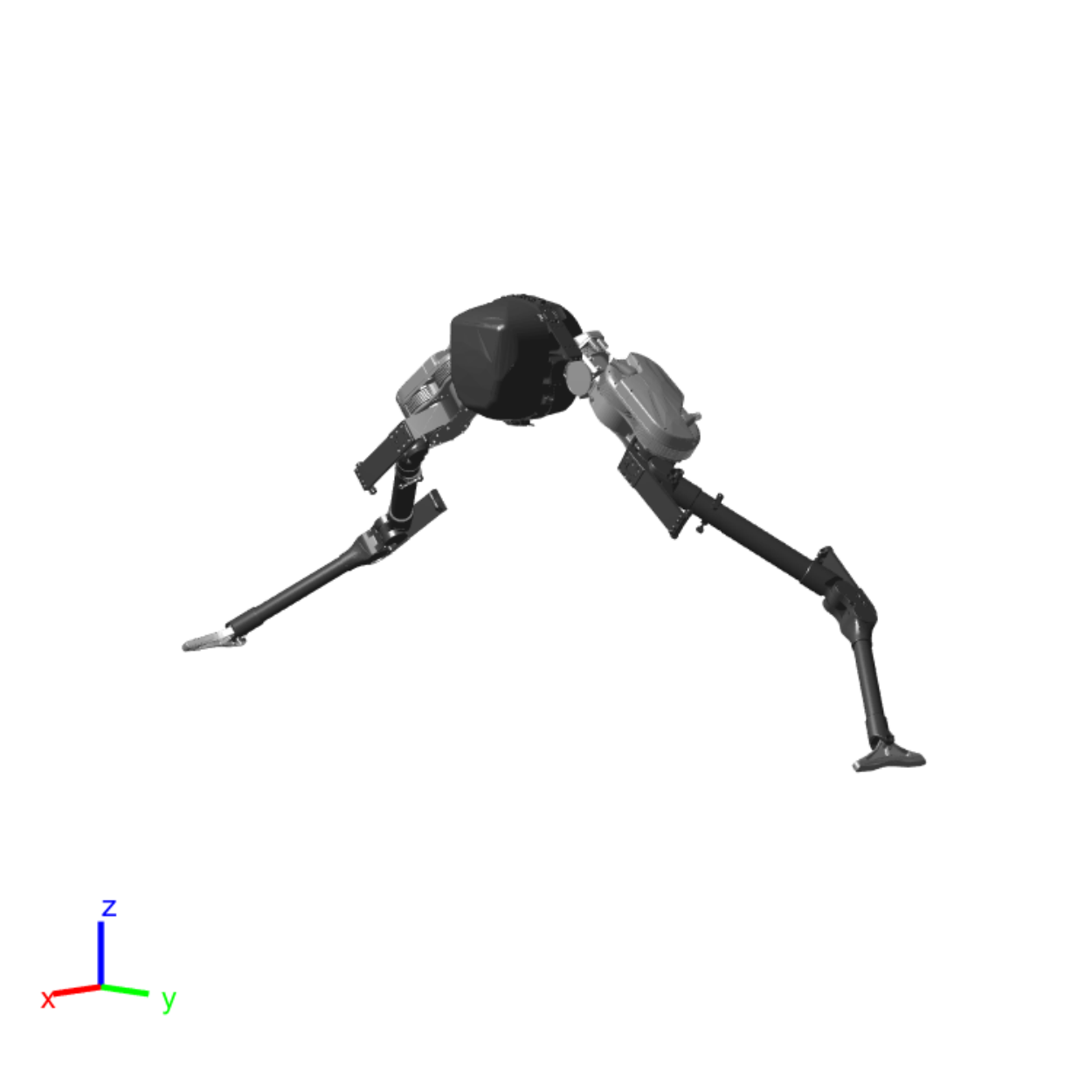} 
        &
        \adjincludegraphics[trim={{0.15\width} {0.166\width} {0.15\width} {0.135\width}},clip,width = .15\linewidth,valign=b]{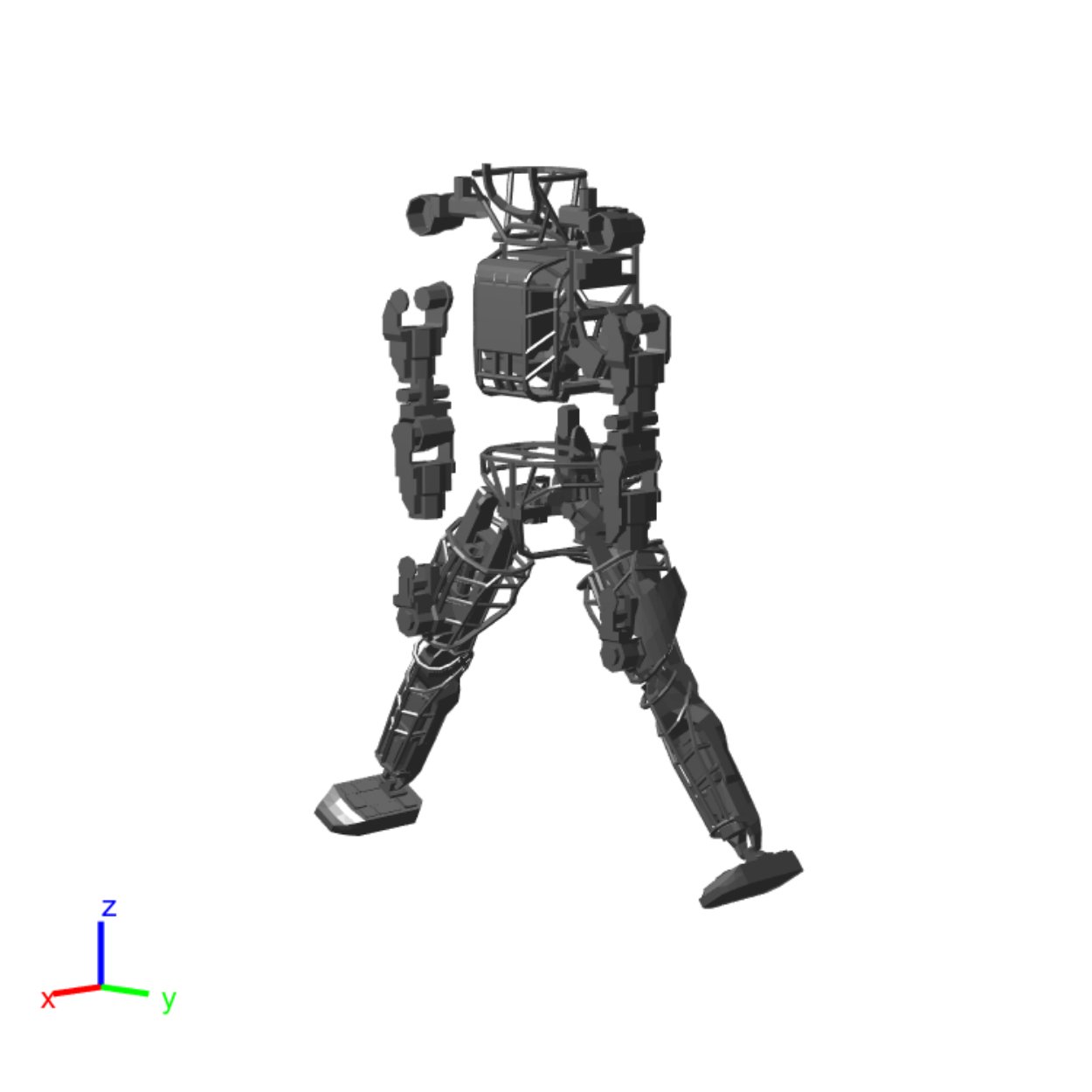} 
        &
        \adjincludegraphics[trim={{0.23\width} {0.15\width} {0.17\width} {0.25\width}},clip,width = .15\linewidth,valign=b]{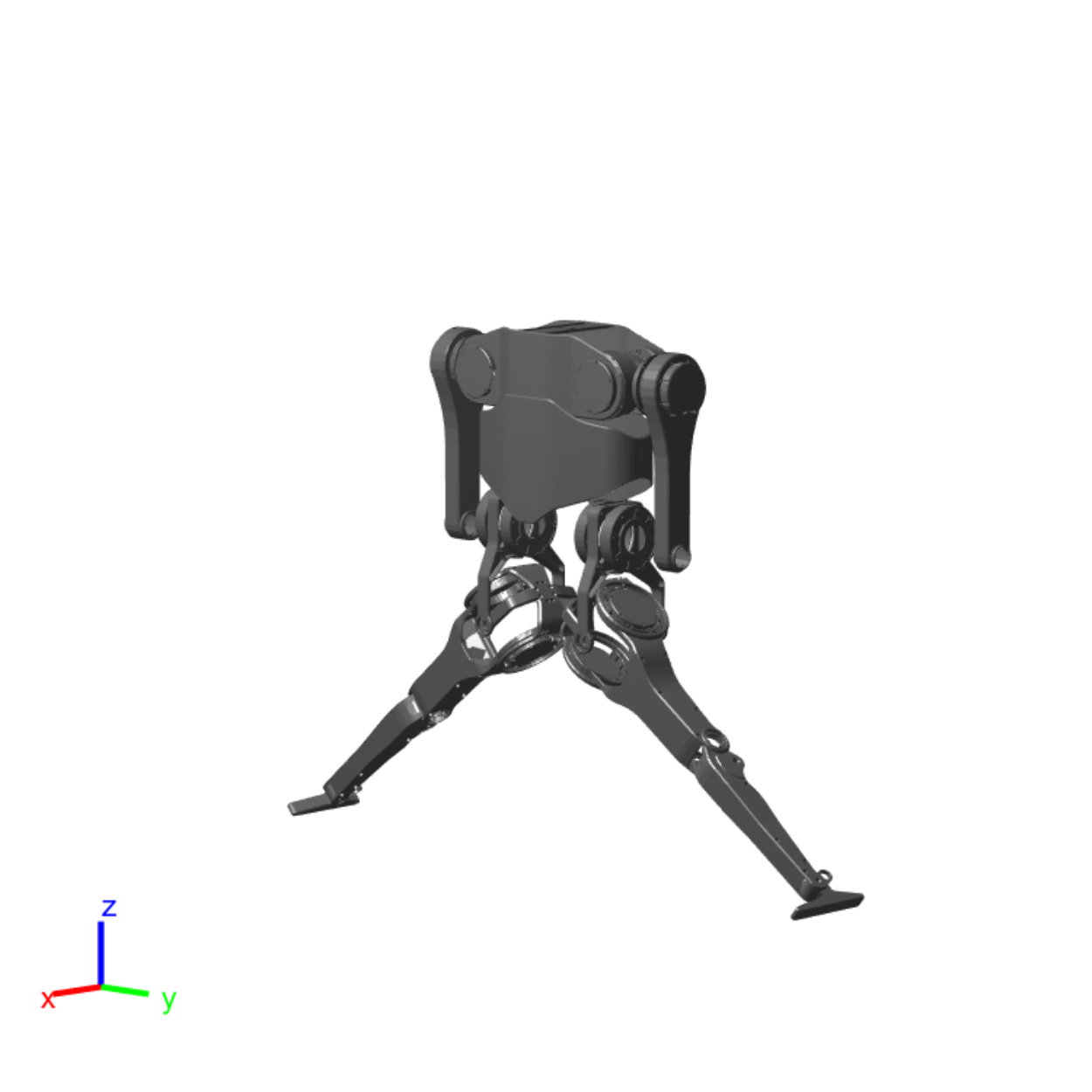} 
        &
        \adjincludegraphics[trim={{0.2\width} {0.2\width} {0.2\width} {0.2\width}},clip,width = .15\linewidth,valign=b]{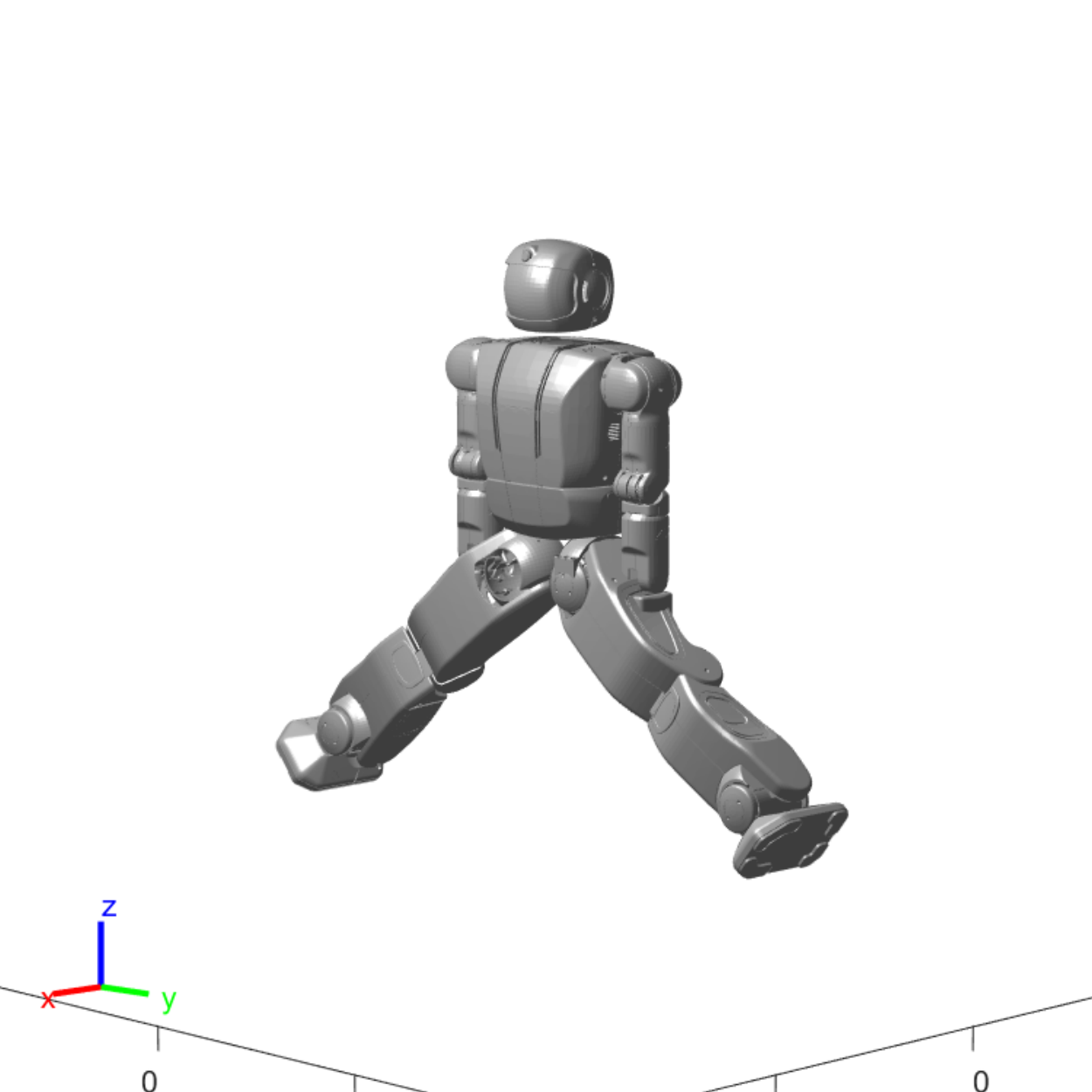} 
        \\
        \bottomrule
    \end{tabular}
    \end{center}
    \caption{Various robots at configurations where CII is maximum or minimum. }
    \label{fig:MinMaxCiiVisualized}
\end{figure}

\begin{figure}[!t]
    \includegraphics[width = .98\linewidth]{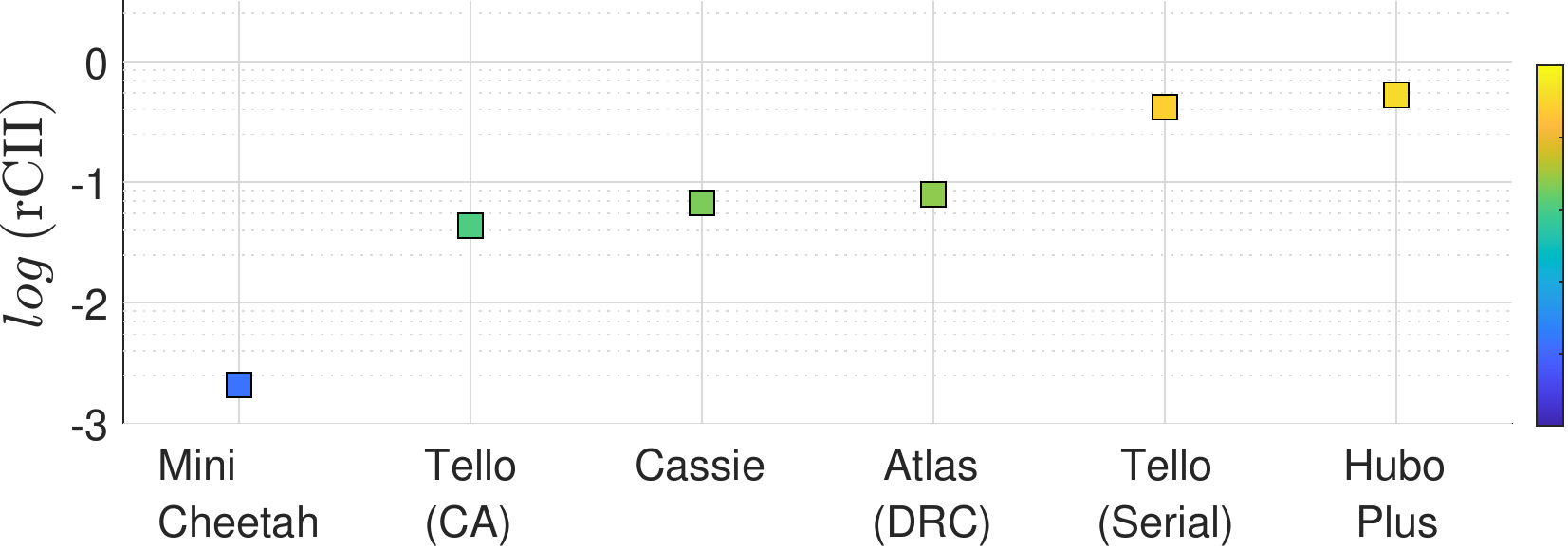}\hspace{-3mm}
    \caption{The logarithm plot of rCII reveals the difference in robot types (e.g. humanoid vs quadruped) and groups (e.g. proximally actuated humanoids vs humanoids whose actuators are near the joint). }
    \label{fig:LogCii}
\end{figure}

One notable characteristic of CII is that CII is a local index; CII is evaluated per a joint configuration. Hence, the isotropy in CCRBI is measured by the distribution of CII evaluated over a set of configurations. In this study, CII of various robots such as quadruped and humanoids that are built with different paradigms are evaluated. Humanoid robots, Tello (CA) and Tello (Serial), are included in this study. The former has all actuators placed around hip joint which will be discussed in section \ref{sec:TelloImplementation}, and the latter is identical to the former except the actuator placement; actuators are placed 
coaxially with its joints. 
The nominal configuration is defined as straight-leg and straight-up pose. The set of test configurations $\bm Q \in \R^2$ was populated by changing two variables: hip abduction/adduction (HAA) angle and hip flexion/extension (HFE) angle. The knee joint angle is chosen such that ankle joint stays on the frontal plane. As a preliminary result, Fig. \ref{fig:CiiDistribution} displays the distribution of $30 \times 30$ equidistance 2D grid of $\textrm{HAA} \in [-\tfrac{\pi}4, \tfrac{\pi}{4}]$, $\textrm{HFE} \in [-\tfrac{\pi}3, 0]$. 

A few notable observations are, firstly, the range of CII (rCII) shows order-difference across different groups of robots. The rCII is defined as follows,
\begin{IEEEeqnarray}{rCl}
    \bar{\bm q} = \argmax_{\bm q} \CII(\bm q, {\bm q}_0), \quad \forall \bm q \in \bm Q, \notag
    \\
    \ubar{\bm q} = \argmin_{\bm q} \CII(\bm q, {\bm q}_0), \quad  \forall \bm q \in \bm Q, 
    \\
    \rCII \coloneqq \CII(\bar{\bm q}, {\bm q}_0) - \CII(\ubar{\bm q}, {\bm q}_0), \notag
\end{IEEEeqnarray}
where $\bm Q$ is a set of sample configurations, ${\bm q}_0$ is a nominal configuration, and $\bar{\bm q}$ and $\ubar{\bm q}$ are configurations where CII are maximum and minimum.
The group of robots designed from the principle of proximal actuation (e.g. MIT Mini Cheetah, Tello (CA), Cassie, Atlas (DRC)) is easily differentiated from a group of robots whose actuators are placed near corresponding joints (e.g. Tello (Serial) and HuboPlus). Within the proximal actuation group as in Fig. \ref{fig:CiiDistProximal}, the humanoids show significantly larger rCII compared to the quadruped. The difference in rCII and the idea of differentiating a class of robot is more obviously demonstrated by taking the logarithm of rCII as in Fig. \ref{fig:LogCii}. Second, visually comparing the configurations where CII is maximum or minimum (Fig \ref{fig:MinMaxCiiVisualized}), a clear trend in posture were observed across robots. With minimum CII configuration, The robots tend to extend their limbs as much as possible, while they crouch to fit inside a ball with maximum CII configuration.

\section{Implementation of Tello Leg} \label{sec:TelloImplementation}

\subsection{Cooperative Actuation for Hip, Knee and Ankle Joints}
A 5-DoF humanoid leg, called Tello, is designed to maximize joint torque with minimal reflected inertia. The actuation system of the Tello leg was built upon the idea of cooperative actuation discussed in \ref{sec:CA}. The core concept of CA is that, if joint torques are not recruited simultaneously, it is beneficial to share loading on a joint with multiple actuators such that torque requirement on the actuators are reduced. In this way, the reflected inertia is rendered smaller, in comparison to employing serial configuration with higher gearing ratio. 
The 2-DoF cooperative actuation is realized in the knee and ankle joint pair.
Specifically speaking of bipedal locomotion, multiple studies state that joint torque of the knee and ankle do not peak simultaneously during nominal walking in case of healthy humans \cite{PATRIARCO1981513, VAUGHAN1996423}, humans with lower limb prosthesis \cite{act3010001}], simulated gaits of bipedal robots \cite{huang2001planning, felis2015gait}. This pattern provides basis to the idea of dynamically shifting the concentration of actuators' torque onto a single joint when needed.

Another case of 2-DoF cooperative actuation is employed in the control of hip abduction and adduction (HAA) and hip flexion and extension (HFE) to promote space optimization to place actuators closer to the center of mass and to double the maximum joint torque at the cost of doubled reflected inertia. 

\subsection{Transmission for 5-DoF Proximal Actuation}
\begin{figure}[!t]
     \centering
     \subfloat[]{
        \label{fig:StickFigure}
        \includegraphics[height=.65\linewidth]{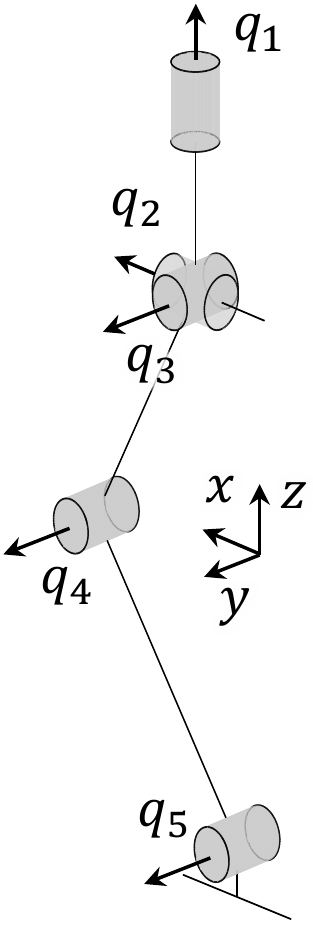}}
        \hspace{15pt}
    \subfloat[]{
        \label{fig:MotorPlacement}
        \includegraphics[height=.65\linewidth]{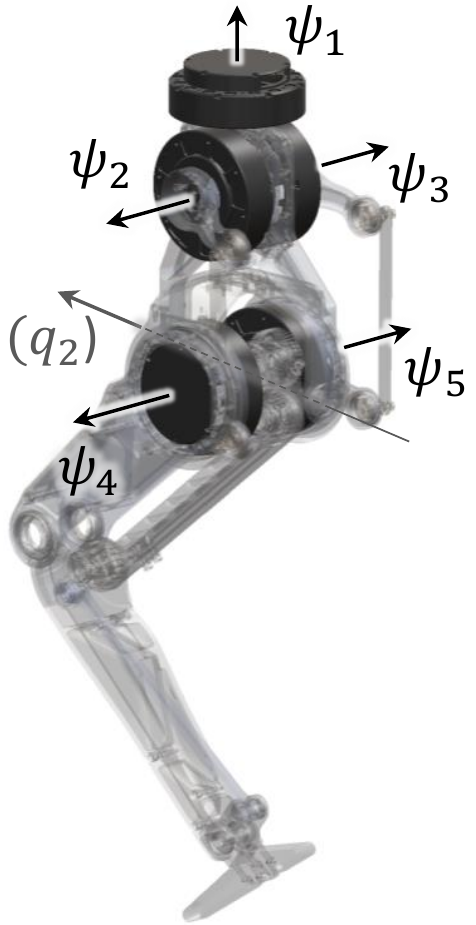}}
    \subfloat[]{
        \label{fig:ParallelMechanism}
        \includegraphics[height=.65\linewidth]{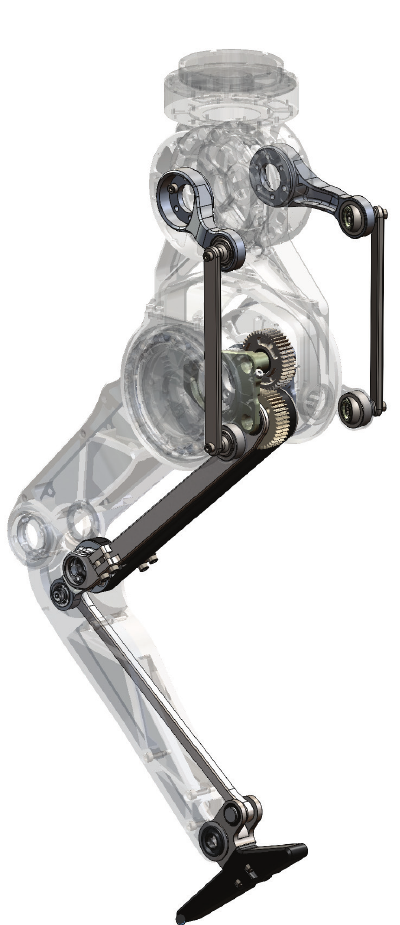}}
    \caption{(a) Stick figure of joint connection (b) Placement of 5 motors (c) Components that complete the parallel loops of actuation topology}
    \label{fig:transmission}
\end{figure}

Towards {maximizing CII}, all 5 motors of Tello are placed around or above its hip joint, as depicted in Fig. \ref{fig:transmission}(b) and their motion and torques are transmitted through a series of mechanisms. For a humanoids leg, Tello leg exhibits comparable CII to other recent humanoids as in Fig. \ref{fig:CiiDistribution}(b). 


The first cooperative actuation is established between a pair of joints and a pair of motors; knee and ankle joint represented as $q_4, q_5$ in Fig. \ref{fig:transmission}(a) and actuator 4 \& 5 whose axes are labed as $\psi_4, \psi_5$ in Fig. \ref{fig:transmission}(b). The movement of the two motors are distributed to two routes by a spur gear differential. One route connects to knee actuation via hamstring-like parallel mechanism in Fig. \ref{fig:transmission}(c) and the other route connects to ankle actuation via a 4-bar linkage located on the shank. The kinematics of knee and ankle differential pair is given as,
\begin{align*}
    q_4 = f_4(\psi_4, \psi_5) = \tfrac12 \psi_4 + \gamma \psi_5 + q_4^o, 
    \\
    q_5 = f_5(\psi_4, \psi_5) = \tfrac12 \psi_4 - \gamma \psi_5 + q_5^o, 
\end{align*}
where $\gamma$ is a geometric parameter of the spur gear differential whose value is $\nicefrac{1}{2}$ in this particular implementation and $q_{(\cdot)}^o$ are joint offsets. 

The second cooperative actuation is installed between another pair of joints and actuators; HAA $q_2$ and HFE $q_3$ as the joint pair and actuator 2 $\psi_2$ and actuator 3 $\psi_3$ as the actuator pair. Note that actuators are placed remotely from the joints of actuation as in Fig. \ref{fig:transmission}(b). The kinematics of this cooperative actuation is nonlinear because 4-bar linkages that connect the motors to joints is not planar. Hence, for the control, the kinematics is numerically approximated by fifth-order polynomials $f_2$ and $f_3$,
\begin{equation*}
    q_2 = f_2(\psi_2, \psi_3), \quad \quad
    q_3 = f_3(\psi_2, \psi_3).
\end{equation*}

Finally, the hip rotation $q_1$ is directly driven by actuator 1 $\psi_1$, which gives, $q_1 = \psi_1$.

The full topology Jacobian $\tensor*[^q]{\bM J}{_\psi} \in \mathbb{R}^{5 \times 5}$ from motor velocity ${\bm v}_\psi \in \mathbb{R}^5$ to joint velocity ${\bm v}_\psi  \in \mathbb{R}^5$ is obtained by taking the derivative of above kinematics, $\bm q = \bm f(\bm \psi)$, 
\begin{gather}
    \tensor*[^q]{\bM J}{_\psi} = \frac{d\bm f}{d \bm \psi}, 
    \\
    \tensor*[^q]{\bM J}{_\psi} \big\vert_{\bm q=0} = \diag\bigg( 1, \bmat{\beta & \beta \\ \nicefrac12 & \nicefrac12 }, \bmat{\nicefrac12 & \vphantom{-}\gamma \\ \nicefrac12 & -\gamma}\bigg),
\end{gather}
where $\tensor*[^q]{\bM J}{_\psi} \vert_{\bm q = \bm 0} $ is an example of linearized actuation topology around home position ($\bm q=\bm 0$) and $\beta=0.465$ is a geometric parameter of the non-planar 4-bar linkage

\section{Experiments} \label{sec:experiments}

This section shows experimental results of a jump achieved with Tello leg. The goal of this experiment is to validate i) the CA embodied for the actuation of knee and ankle joint and ii) transmission design for minimizing CII of 5-DoF leg. 
The Tello leg affixed at the end of a passive 2-DoF boom which allows rotation in elevation and azimuth. The leg was placed on a force place to measure the ground reaction force (GRF) at 1kHz. For this particular experiment, the first joint was locked and foot was removed so that the ankle joint makes contact with the force plate.

\subsection{Hardware Setup}

A real-time controller (National Instruments, sbRIO-9627) reads displacements of rotors of 5 actuators and estimates joint velocities with quadrature encoders (AMS, AS5147U) via digital I/O and commands reference torques to BLDC motor drivers (Advanced Motion Control, FM060-CM-25) via analog channel. This control loop runs at 1kHz. Tello is equipped with quasi-direct drive motors (Steadywin, V3) which contain a single-stage 6:1 planetary gearbox. This actuator has similar properties to those of the Mini Cheetah \cite{MiniCheetah}. The mass of the entire setup including the robot and the boom measures $m_t=6.9$ kg. 

\subsection{Control for jumping}
A state machine administers jumping and landing in following order of state transitions. In the beginning, the Tello maintains \textit{Ground} state where impedance controller keeps the knee bent, so that it can accumulate enough momentum during thrust to make a jump. 
When a jump is commanded, the state machine is in \textit{Thrust} state where pure feedforward force control is employed. The feedforward force ${\bm \tau}_x$ has z-directional component only which is prescribed by a fifth-order B\'ezier curve. Next, when vertical GRF measured from the force plate reaches lower threshold ($8$ N), the state changes to \textit{Aerial} state where joint positional and differential (PD) control is engaged to retract the foot to prevent falling into singularity (straight leg) and prepare soft landing. Next, upon detecting increase in GRF due to contact, the state turns back into \textit{Ground}. To prevent bouncing between states due to unstable contact, the \textit{Ground} state is locked for $2$ s after the first entry from the \textit{Aerial} state to the \textit{Ground} state.  

The control law for actuator torques ${\bm \tau}_\psi \in \mathbb{R}^5$ per state writes 
\begin{equation}
    {\bm{\tau}}_\psi \!
    = \!
    \begin{cases}
        \tensor*[^x]{\bM J}{_\psi^\top}
        \big( 
            m_t\bm  g 
            +
            {\bM K}_x^p \delta \bm x
            + 
            {\bM K}_x^d \delta \dot{\bm x}
        \big ) \!\!\!\!\!
        &(\textrm{stance}) 
        \\[5pt]
        \tensor*[^x]{\bM J}{_\psi^\top} \bm \tau_x
        &(\textrm{thrust})
        \\[5pt]
        \tensor*[^q]{\bM J}{^\top_\psi}
        \big(
            \tensor*[]{\bM K}{^p_q} \delta \bm q 
            +
            \tensor*[]{\bM K}{^d_q} \delta \dot{\bm q} 
        \big)
        &(\textrm{flight})
    \end{cases} 
\end{equation}
where the dimension of Jacobians are $\tensor*[^x]{\bM J}{_\psi^\top}, \tensor*[^q]{\bM J}{_\psi^\top} \in \R^{5\times 3}$, $\bm g \in \R^3 $ is the gravitational acceleration, ${\bM K}_x^p, {\bM K}_x^d \in \mathbb{R}^{3\times3} $ are task space PD gain matrices, $\delta {\bm x}, \delta \dot{{\bm x}}$ are position and velocity error in task space, ${\bM K}_q^p, {\bM K}_q^d \in \mathbb{R}^{5\times 5}$ are joint space PD gain matrices, $\delta {\bm q}, \delta \dot{{\bm q}}$ are position and velocity error in joint space. 
Next, the individual entries of actuator torque vector ${\bm \tau}_\psi$ are saturated to $[10, -10]$ Nm in software for safety {($\approx67\%$ peak motor torque)}.

\begin{figure}[t]
    \centering
    \includegraphics[width = 0.97\linewidth]{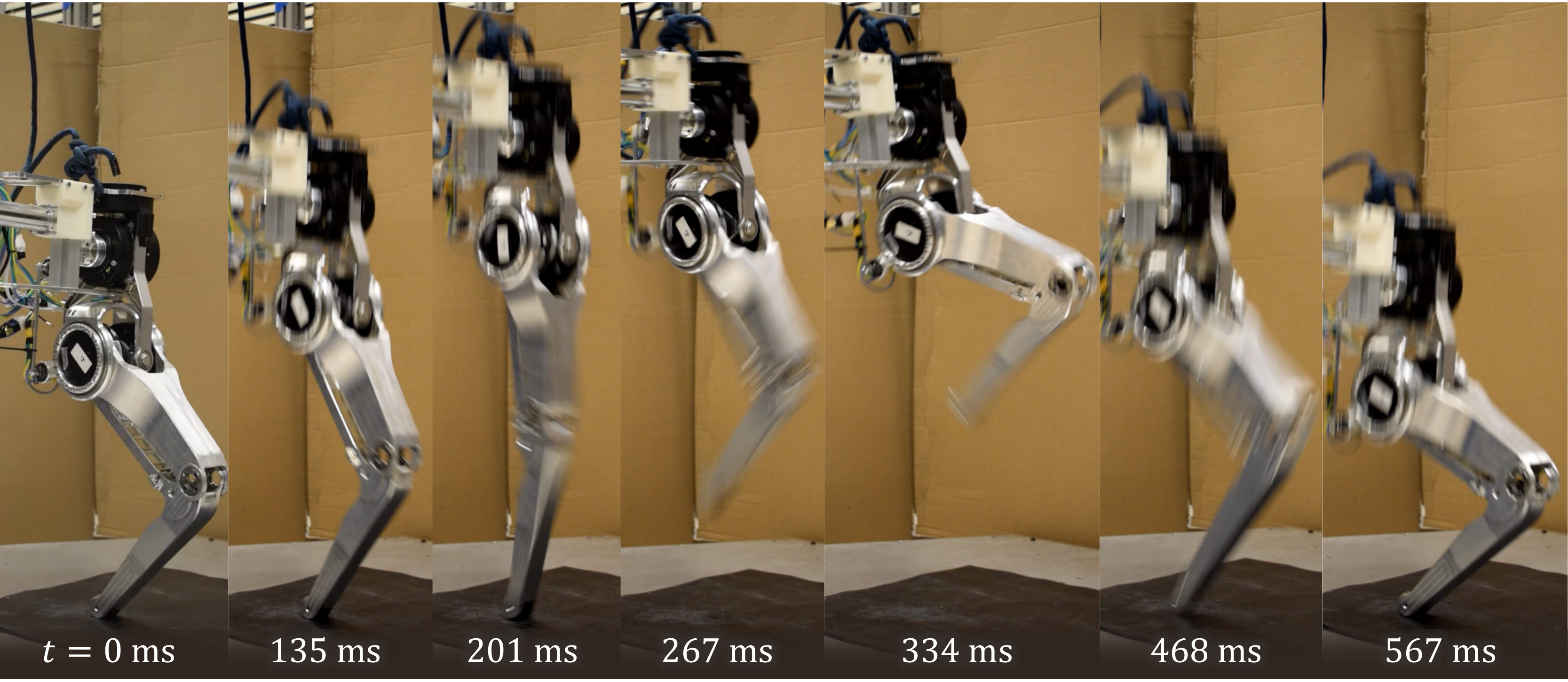}
    \caption{Jumping in place. With the design philosophy and realization of transmission described in this paper, the robot weighing $6.9$ kg jumped successfully with 5 motors attached around its hip.} \label{fig:TelloJump}
\end{figure}
\begin{figure}[!t]
    \centering
    \includegraphics[width = 0.97\linewidth]{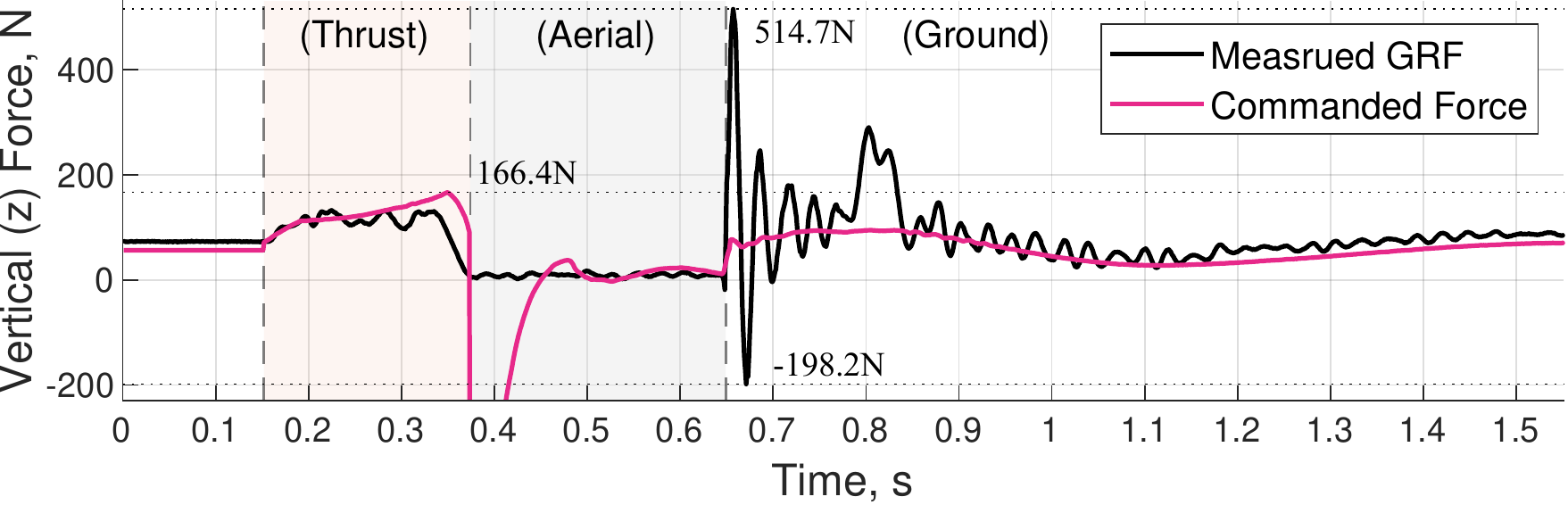}
    \caption{The vertical GRF measured from a force plate and commanded task space force. The impact due to landing incurred vibration of the force plate which resulted in oscillation in GRF measurement. } \label{fig:jumpTaskSpace}
\end{figure}

\begin{figure*}[!t]
     \centering
     \subfloat[]{
        \label{fig:torque_velocity_KneeAnkle}
        \includegraphics[width = 0.4559\linewidth]{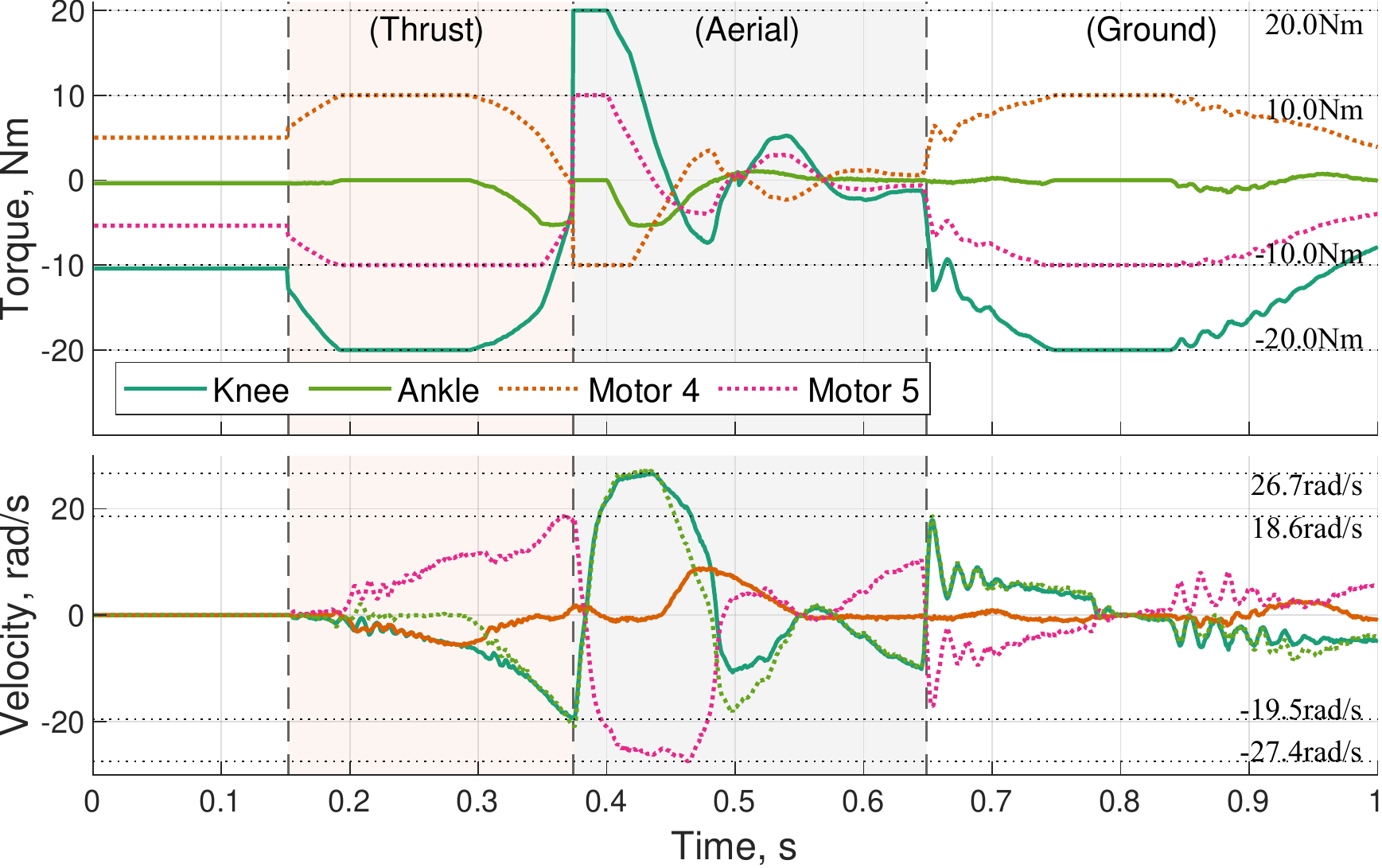}
    }
    \subfloat[]{
        \label{fig:TCP_VCP_KneeAnkle}
        \adjincludegraphics[trim={{0.09\width} {0.31\height} {0.10\width} {0.33\height}},clip,width = .5044\linewidth]{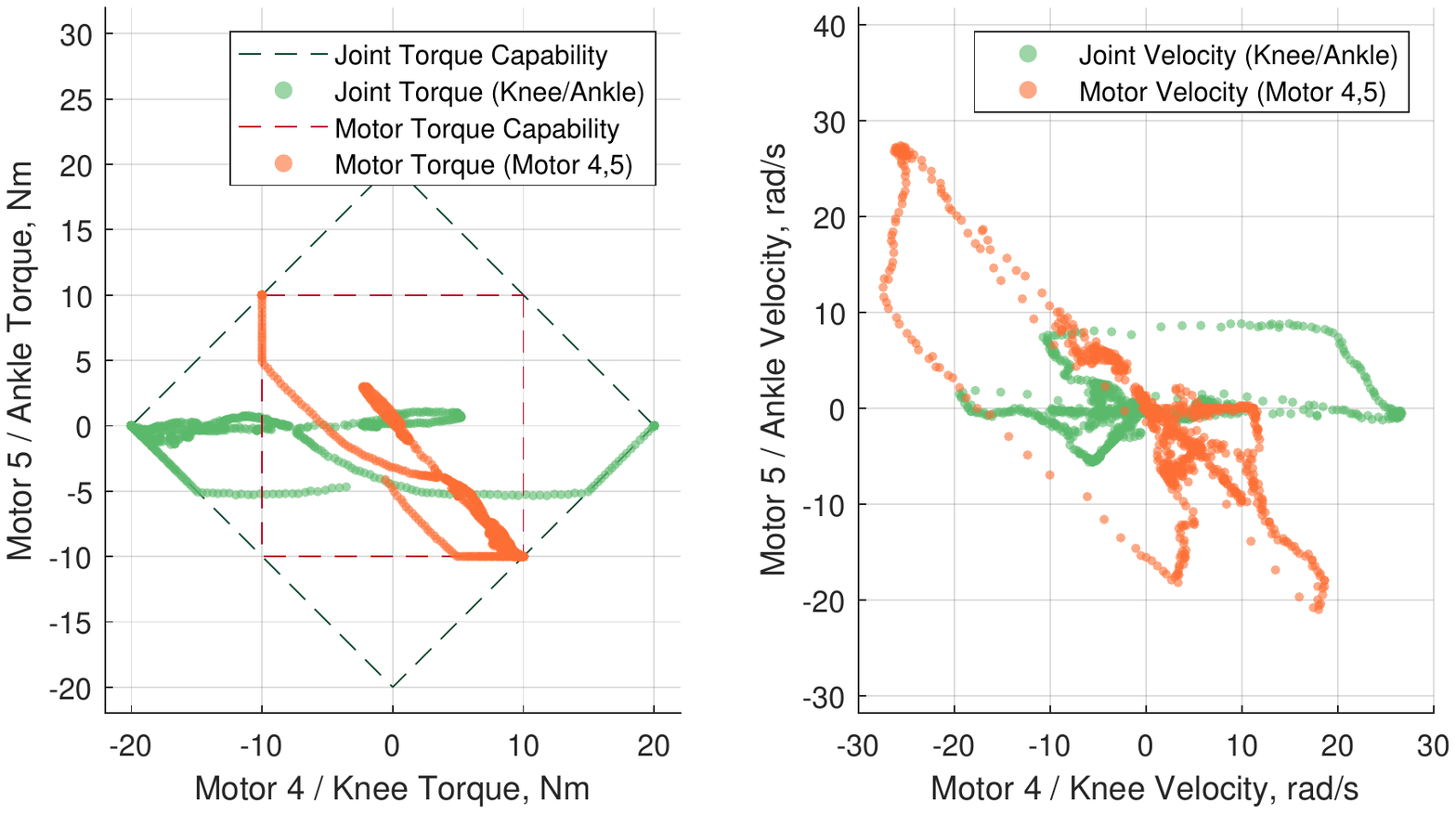}
    }
    \caption{(a) The trajectories of torques and velocity of a CA pair involving ankle AFE, APDF, actuator 4 and 5. (b) The trajectories of torque (left) and velocity (right) of KFE \& APDF joints and actuators 4 \& 5. }
    \label{fig:ResultKneeAnkleCA}
\end{figure*}

\begin{figure*}[!t]
     \centering
     \subfloat[]{
        \label{fig:torque_velocity_Hip}
        \includegraphics[width = 0.4559\linewidth]{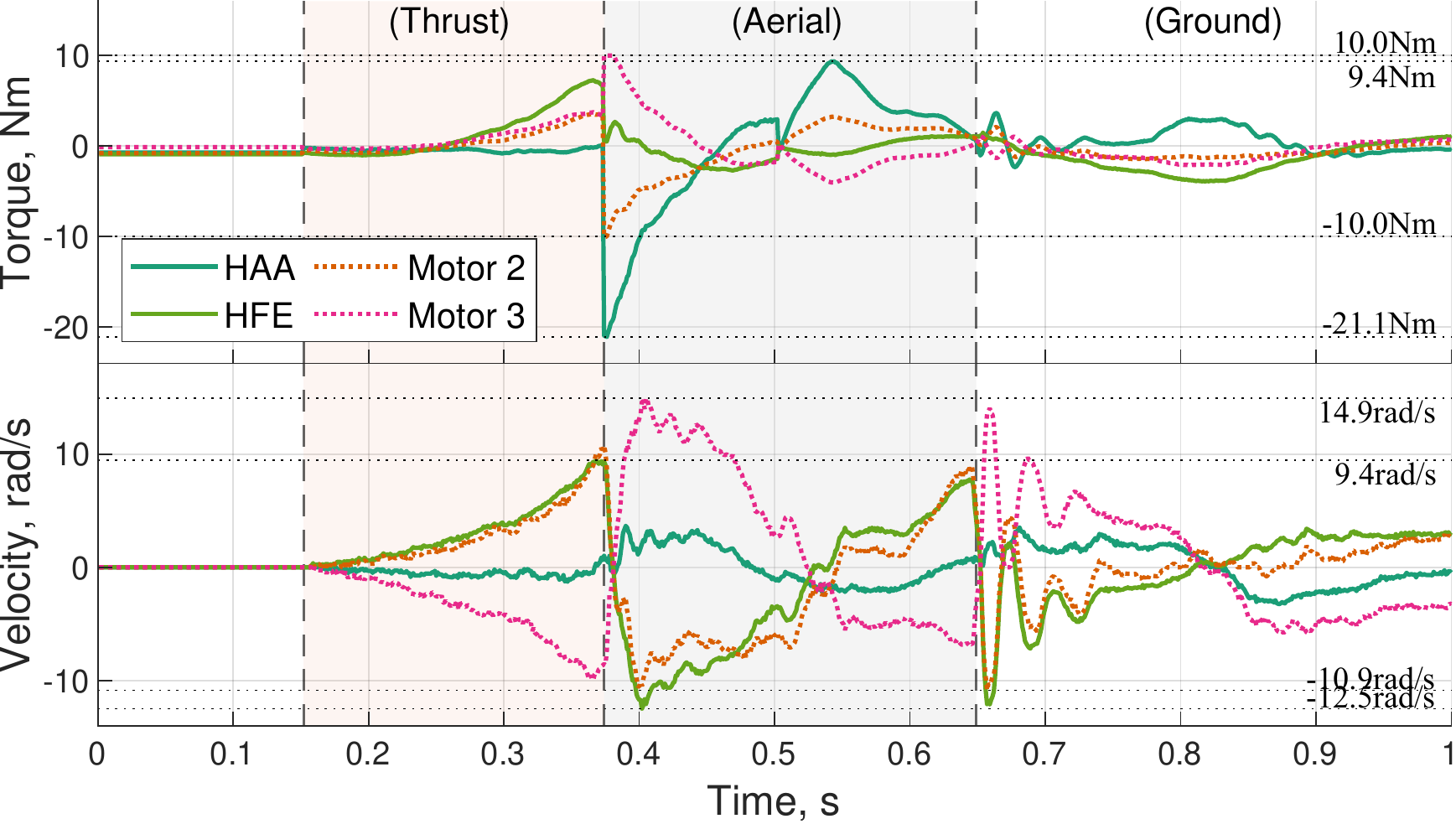}
    }
    \subfloat[]{
        \label{fig:TCP_VCP_Hip}
        \adjincludegraphics[trim={{0.09\width} {0.31\height} {0.10\width} {0.33\height}},clip,width = .5044\linewidth]{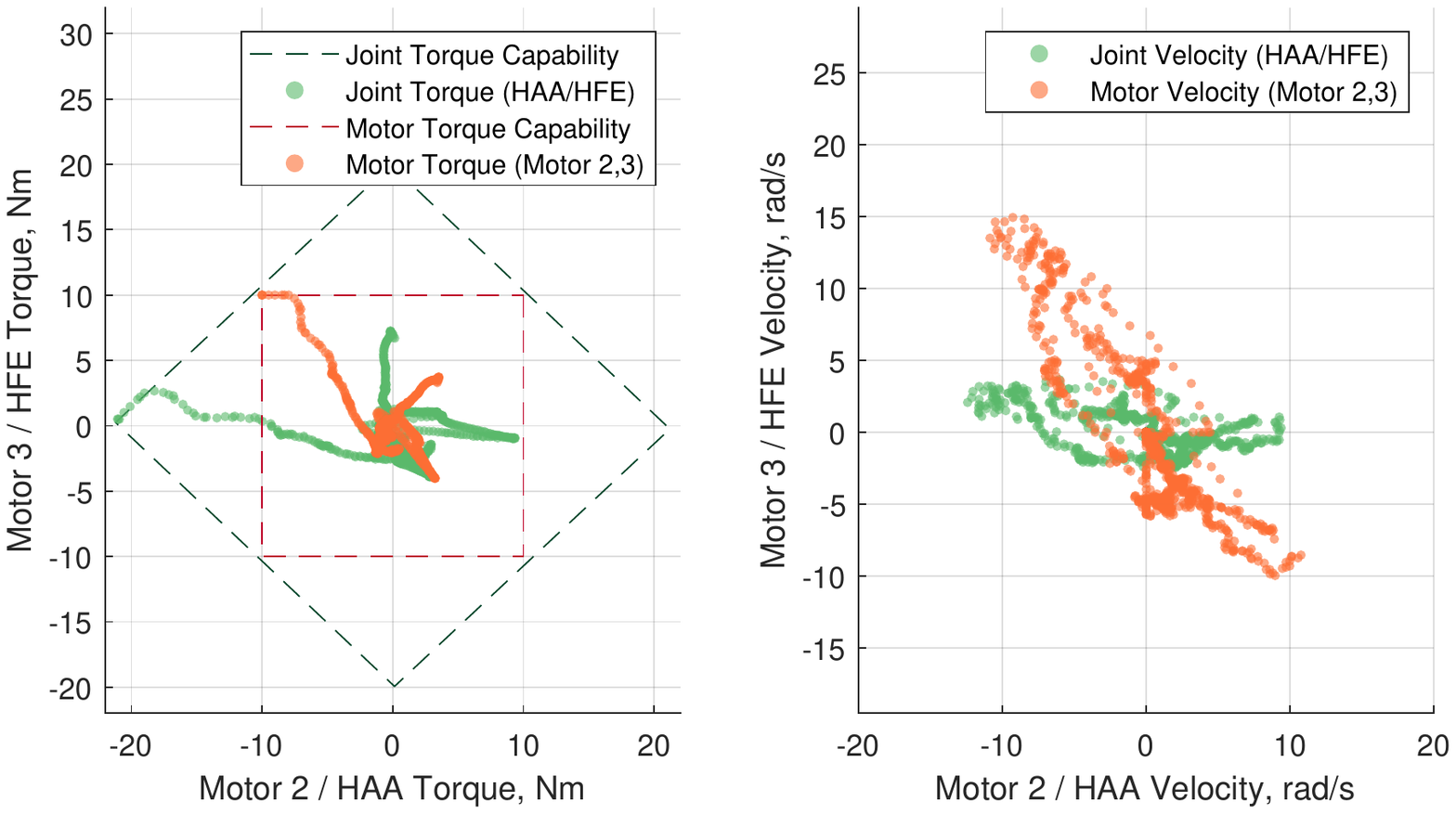}
    }
    \caption{(a) The trajectories of torques and velocity of a CA pair involving HAA, HFE, actuator 2 and 3.  (b) The trajectories of torque (left) and velocity (right) of HAA \& HFE joints and actuators 2 \& 3. }
    \label{fig:ResultHipCA}
\end{figure*}

\subsection{Results}
The Tello leg successfully jumped $171$ mm vertically and landed softly as in Fig. \ref{fig:TelloJump}. The $Thrust$ state took $220$ ms and stayed in $Aerial$ state for $275$ s. 

Figure \ref{fig:jumpTaskSpace} compares vertical GRF measured by the force plate and the commanded task space force (or torques projected onto task space). In most time throughout the $Thrust$ state, the measusred GRF aligns with the commanded force. The reasons to the mismatch are i) the controller does not consider Coulomb and viscous friction inertial forces to accelerate limbs and ii) Second, as the leg extends to make a jump, the angular velocity of actuators soars up to maximally $26.7$ rad/s or minimally $-27.4$ rad/s as shown in Fig. \ref{fig:ResultHipCA} (a) which can limit the maximum torque applied. At the beginning of the \textit{Aerial} state, joint space PD controller requires maximum joint torque to retreat the leg to slightly-bent-knee pose. At this moment, the knee is almost straight or the manipulator Jacobian is near singularity. Hence, though the projected force on task space is rendered significantly large (plunge in commanded force at 0.37 s mark in Fig. \ref{fig:jumpTaskSpace}).
The high frequency ringing in vertical GRF reading after $0.66$ s mark Fig. \ref{fig:jumpTaskSpace} is due to oscillation of the force plate incurred by landing impact.

The effectiveness of CA, which is to utilize differential configuration to channel actuator torques on to a joint at a time is verified. 
With actuators whose maximum torque is $\pm 10 $ Nm, larger knee joint torque, $-20$ Nm, was observed during the $Thrust$ state as in Fig. \ref{fig:ResultKneeAnkleCA}(a) which is critical for explosive jumping. Then $20$ Nm on knee joint was observed the onset of the $Aerial$ state which was to retract the foot. 
The true benefit of utilizing CA on the knee and ankle joint pair is revealed in Fig. \ref{fig:ResultKneeAnkleCA}(b) which is to create large knee joint torque by combining two actuator torques. The ankle torque was observed relatively smaller than the knee torque, which aligns with the fact that ideal vertical jump does not involve substantial ankle torque.

Fig. \ref{fig:ResultHipCA} displays the torques and velocities observed in the other differential pair involving HAA \& HFE joint and actuator 2 \& 3. The HFE torque which is essential for pushing the ground increases towards the end of $Thrust$ state. At the instance of take-off, the ankle slid ~$10 $ cm in medio-lateral direction (y-axis) because the boom generates spherical motion while the leg tries to jump vertically. As a consequence, at the beginning of $Aerial$ state, joint space PD controller immediately employs maximum torques of actuators 2 \& 3 to pull the ankle joint back to the saggital plane by applying large HAA torque of $-21.1$ Nm. 

In contrast to the fact that actuator torques can be added up to create larger joint torque with differential configuration, twice larger velocity than joint velocity may be requested on actuators, theoretically. However, in this experiment, actuator velocities were not necessarily twice larger than joint velocities. 
As in Fig. \ref{fig:ResultKneeAnkleCA}(a), the CA employed in knee and ankle joint showed actuator velocities ranging within $[-27.4, 26.7]$ rad/s, while joint velocities were in $[-19.5, 26.7]$ rad/s. In case of CA employed in HFE \& HFE joints (Fig. \ref{fig:ResultHipCA}(a)), the actuator velocities were within $[-10.9, 14.9]$ rad/s, while joint velocities ranged within $[-12.5, 9.4]$ rad/s. 

\subsection{Discussion}
The jump experiment demonstrated that CA employed in the knee and ankle joint pair can recruit two actuators to exert twice larger torque $20$ Nm on knee joint than maximum actuator torque $10$ Nm. At the same time, the ankle torque was kept small ($<5$ Nm) such that majority of the torque outputs of the two actuators are routed onto the knee joint. 
Towards minimizing CII, the Tello leg has all five actuators placed near the center of mass. The second case of cooperative actuation which is the differential configuration on the HAA and HFE joint helped optimizing the actuators' placement. The actuators 2 \& 3 were placed remotely from the joint such that it allows actuators 4 \& 5 which drives knee and ankle joints can be placed on the hip joint. 
One concern was that differential configuration may require actuators to run faster than joint velocities (e.g. twice faster, maximally). {However, the jump experiment showed that the ranges of actuator velocity was not necessarily twice larger than the ranges of joint velocity, they are rather similar.}
One downside of the transmission design for 5-DoF proximal actuation (Fig. \ref{fig:transmission} is the complexity of the mechanical build. For instance, the range of motion of ankle flexion is limited due to the chain of 4-bar linkages that connects actuators placed on the hip to the ankle joint. 
Still, the CA for ankle and knee operation has shown great benefit. Hence, to counter the drawback of increased mechanical complexity, other space optimization methods and more efficient design choices for parallel transmission will be explored in future studies.
Although substantial design effort was invested in the build of Tello, the resultant rCII (Fig. \ref{fig:LogCii}) is only slightly better than Cassie, which employs one actuator attached closer to the distal side or right under the tarsus joint. Therefore, authors will investigate in future work if the the transmission design of Tello guarantees ultimately simpler control or superior performance envelope than a robot like Cassie.

\section{Conclusion}
The proposed design principle of cooperative actuation which aims to generate large force in bipeds while minimizing reflected inertia so that legged robots can rapidly regulate interaction with environment. We also propose a performance metric CII which quantifies the degree to proximal actuation, or how much mass is concentrated toward the center of mass. The CII was evaluated for different types and class of existing robots and it successfully demonstrated that CII can predict the type of robot and the placement of actuators of a robot. We have built a 5-DoF leg for humanoid robot, Tello, upon the idea of CA. All 5 actuators of the leg are placed near the center of mass to promote more efficienct control synthesis and implementation. The Tello leg was able to jump by recruiting multiple motors to amplify joint torques. 

\ifCLASSOPTIONcaptionsoff
  \newpage
\fi



\bibliographystyle{IEEEtran}
\bibliography{main_2022RAL.bib}
\end{document}